\documentclass[runningheads]{llncs}

\usepackage{eccv}

\usepackage{eccvabbrv}

\usepackage{booktabs}
\usepackage{multirow}
\usepackage{amsmath}
\usepackage[table]{xcolor}
\usepackage{xcolor}
\usepackage{graphicx}
\usepackage{subcaption}     % 子图管理（含subfigure环境）
\usepackage{float}          % 提供[H]选项固定位置
\usepackage{lipsum} 
\usepackage{bm}
\usepackage[title,titletoc]{appendix}
\usepackage{mathtools} 
\usepackage{tikz}
\usepackage{pifont}
\usepackage{amssymb}
\usepackage{wasysym}
\usepackage{algorithm}
\usepackage{algpseudocode}
\usepackage{amsmath}
\usepackage{amsfonts}
\usepackage{color}
\usepackage{colortbl}
\usepackage{wrapfig}

\usepackage[accsupp]{axessibility}  % Improves PDF readability for those with disabilities.

\definecolor{mygray}{gray}{0.3}
\definecolor{lightgray}{gray}{0.95}
\usepackage{hyperref}

\usepackage{orcidlink}

\begin{document}

\title{Trust the Unreliability: \textit{Inward–Backward} Dynamic Unreliability-Driven Coreset Selection for Medical Image Classification} 

\titlerunning{ Coreset Selection for Medical Image Classification}
% \titlerunning{Dynamic Unreliability-Driven Coreset Selection for Medical Image Classification}

\author{Yan Liang\inst{1} \and
Ziyuan Yang\inst{1} \and
Zhuxin Lei\inst{1} \and
Mengyu Sun\inst{1} \and
Yingyu Chen\inst{1} \and
Yi Zhang\inst{1}}

\authorrunning{Liang et al.}

% Springer Heidelberg, Tiergartenstr.~17, 69121 Heidelberg, Germany

\institute{
Sichuan University, Chengdu 610207, P. R. China
}

\maketitle

\begin{abstract}
Efficiently managing and utilizing large-scale medical imaging datasets with limited resources presents significant challenges.
While coreset selection helps reduce computational costs, its effectiveness in medical data remains limited due to inherent complexity, such as large intra-class variation and high inter-class similarity.
To address this, we revisit the training process and observe that neural networks consistently produce stable confidence predictions and better remember samples near class centers in training.
However, concentrating on these samples may complicate the modeling of decision boundaries. Hence, we argue that the more unreliable samples are, in fact, the more informative in helping build the decision boundary.
Based on this, we propose the \textbf{Dynamic Unreliability-Driven Coreset Selection~(DUCS)} strategy. 
Specifically, we introduce an inward-backward unreliability assessment perspective: 
1) Inward Self-Awareness: The model introspects its behavior by analyzing the evolution of confidence during training, thereby quantifying uncertainty of each sample.
2) Backward Memory Tracking: The model reflects on its training tracking by tracking the frequency of forgetting samples, thus evaluating its retention ability for each sample.
Next, we select unreliable samples that exhibit substantial confidence fluctuations and are repeatedly forgotten during training. This selection process ensures that the chosen samples are near the decision boundary, thereby aiding the model in refining the boundary.
Extensive experiments on public medical datasets demonstrate our superior performance compared to state-of-the-art~(SOTA) methods, particularly at high compression rates.
\end{abstract}

\section{Introduction}
\label{sec:intro}
In the field of medical artificial intelligence, efficient management and analysis of medical image data are critical for accurate and reliable computer-aided diagnosis. Medical image datasets are typically very large and complex, which leads to increased storage and transmission costs, as well as heavy computational demands during model training~\cite{yang2025patient}.
Hence, in resource-constrained scenarios, improving computational efficiency while maintaining diagnostic precision remains a significant challenge~\cite{suganyadevi2022review,li2023medical}.

\begin{wrapfigure}{r}{0.5\textwidth}
  \centering
  \includegraphics[width=0.48\textwidth]{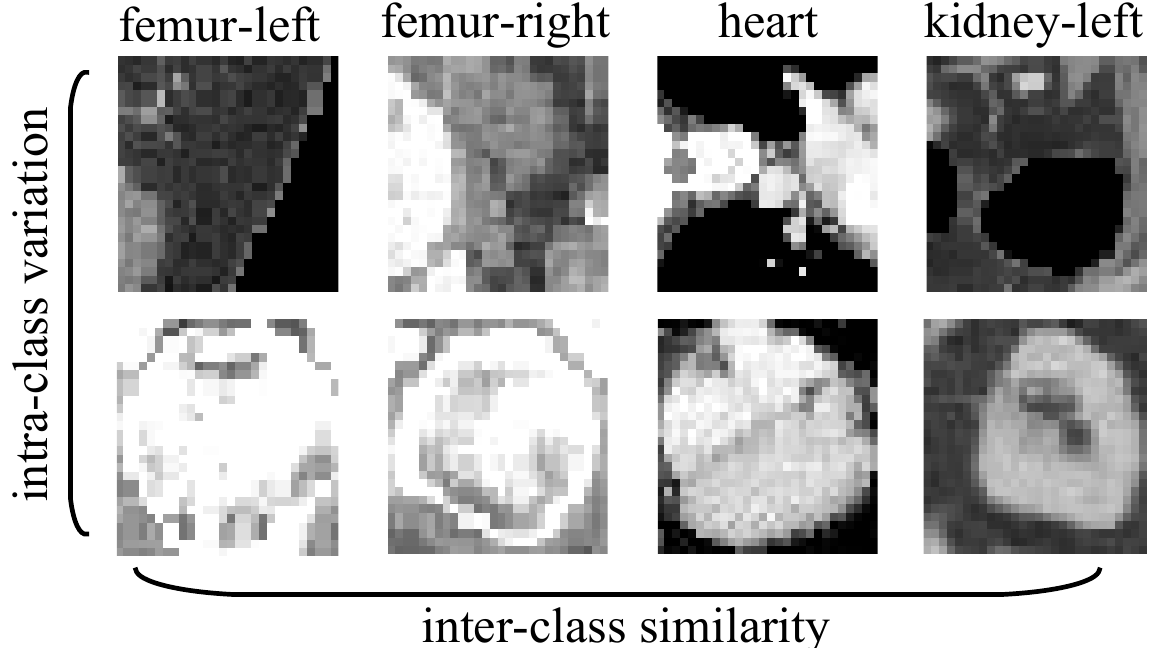}
  \caption{Examples of intra-class variation and inter-class similarity in medical image classification.}
  \label{fig:medical_image}
  \vspace{-20pt}
\end{wrapfigure}
Coreset selection has been widely acknowledged as an effective approach to address these challenges~\cite{huang2023efficient,marion2023less,yang2024mind}. 
Existing approaches primarily assess sample importance through two paradigms consisting of score-based methods which utilize metrics such as classification confidence~\cite{coleman2019selection} and gradient norms~\cite{mirzasoleiman2020coresets,killamsetty2021grad,paul2021deep} as well as geometry-based methods~\cite{sorscher2022beyond} that aim to optimize data coverage by minimizing redundancy among selected samples.
For instance, Pleiss \textit{et al.}~\cite{pleiss2020identifying} measured the probability gap between the target class and the second-highest predicted class at each epoch. Paul \textit{et al.}~\cite{paul2021deep} evaluated the impact of individual samples on parameter updates by computing the L2 norm of their loss gradients. Hong \textit{et al.}~\cite{hong2024evolution} used two windows to monitor the training process and measured changes in sample importance by evaluating the variance of error scores.
% 这里的蓝色是添加的
Zhang \textit{et al.}~\cite{zhang2024spanning} proposed to select samples by evaluating temporal fluctuations in KL-divergence-based loss differences during training, where the observation range $T$ decreases as the pruning rate increases. Moreover, Xia \textit{et al}~\cite{xia2023moderate} selected samples whose distances to their respective k-means cluster centers are closest to the median distance.
% Xia \textit{et al.}~\cite{xia2023moderate} favor those in the vicinity of the median. 
However, medical image data inherently exhibit substantial intra-class variations and high inter-class similarity~\cite{song2015large}, which inevitably blur decision boundaries and thereby limit the effectiveness and generalization of coreset selection methods in this domain, as depicted in Figure.~\ref{fig:medical_image}.

To address these challenges, this paper revisits the training dynamics of deep neural networks. 
We find that neural networks tend to exhibit more stable prediction behavior and enhanced memory retention for samples located in dense regions near class centers. Therefore, overemphasizing these samples may lead the model to overfit to the densest regions, thereby blurring decision boundaries, especially for medical image datasets.
Based on this observation, a counterintuitive yet crucial insight emerges: unreliable samples near decision boundaries carry the highest informational value, effectively revealing the model’s cognitive boundaries and playing a vital role in refining its decision boundaries~\cite{thulasidasan2019mixup,yang2024mind}.
 
In this paper, we present a novel \textit{\textbf{Dynamic Unreliability-Driven Coreset Selection (DUCS)}} method. DUCS quantifies the unreliability of each sample from two perspectives: examining the model's self-awareness of its own predictions (looking inward) and memory tracking (looking backward). 

\textbf{\textit{Inward-Looking Perspective:}} Rather than directly mapping outputs to deterministic class probabilities using softmax, we model the output as a Dirichlet distribution, representing the belief over class probabilities~\cite{jsang2018subjective,sensoy2018evidential,corbiere2021beyond}. 
The softmax function often results in a winner-takes-all effect, which can lead to overfitting when training data is limited. This tendency leads to sharp decision boundaries, which in turn degrades generalization performance.
Unlike the softmax function, the Dirichlet distribution models class beliefs rather than deterministic probabilities. For low-confidence samples, its flatter shape helps mitigate overfitting and yields smoother decision boundaries.
As the model evolves, samples exhibiting greater confidence fluctuations are typically closer to decision boundaries, and thus make them more sensitive to subtle changes in the model.
This sensitivity reflects the model’s self-awareness of its uncertainty, which is leveraged in DUCS to quantify sample unreliability from the inward-looking perspective.

\textbf{\textit{Backward-Looking Perspective:}} During training, due to the stochasticity in optimization and data ordering, the model may forget previously learned samples across different training epochs~\cite{toneva2018empirical}. 
Forgetting occurs most frequently for samples near decision boundaries, which repeatedly undergo ``learning and forgetting" cycles due to their high sensitivity to model updates, as the model struggles to maintain stable representations.
Hence, DUCS directs the model to look backward at its memory tracking and introduce a ``forgetting frequency" metric, which reflects the tracking learning and forgetting over epochs.

To achieve an inward-backward reliable-level quantification method for each sample, we construct a novel unreliability metric that integrates dynamic changes in the model's confidence and the forgetting frequency throughout the training process.

This metric enables the systematic assessment of the potential information value of samples, providing reliable guidance for the efficient and robust compression of medical image data. Extensive experiments on several medical image datasets demonstrate that the proposed DUCS achieves superior performance compared to state-of-the-art~(SOTA) methods, particularly in extremely low data selection scenarios. Our main contributions can be summarized as follows:
\begin{itemize}
\item We propose a novel coreset selection strategy, Dynamic Unreliability-Driven Coreset Selection (DUCS), to efficiently select unreliable yet informative samples.
\item We propose quantifying unreliability from inward- and backward-looking perspectives: the former measures confidence evolution to capture self-awareness, while the latter analyzes memory tracking via forgetting frequency.
\item We develop an unreliability metric, the Unreliability Score, that combines confidence fluctuation and forgetting dynamics, providing systematic guidance for coreset construction and data compression.
\end{itemize}

\label{sec:Method}
\begin{figure*}[t]
    \centering
    \includegraphics[width=1.0\textwidth]{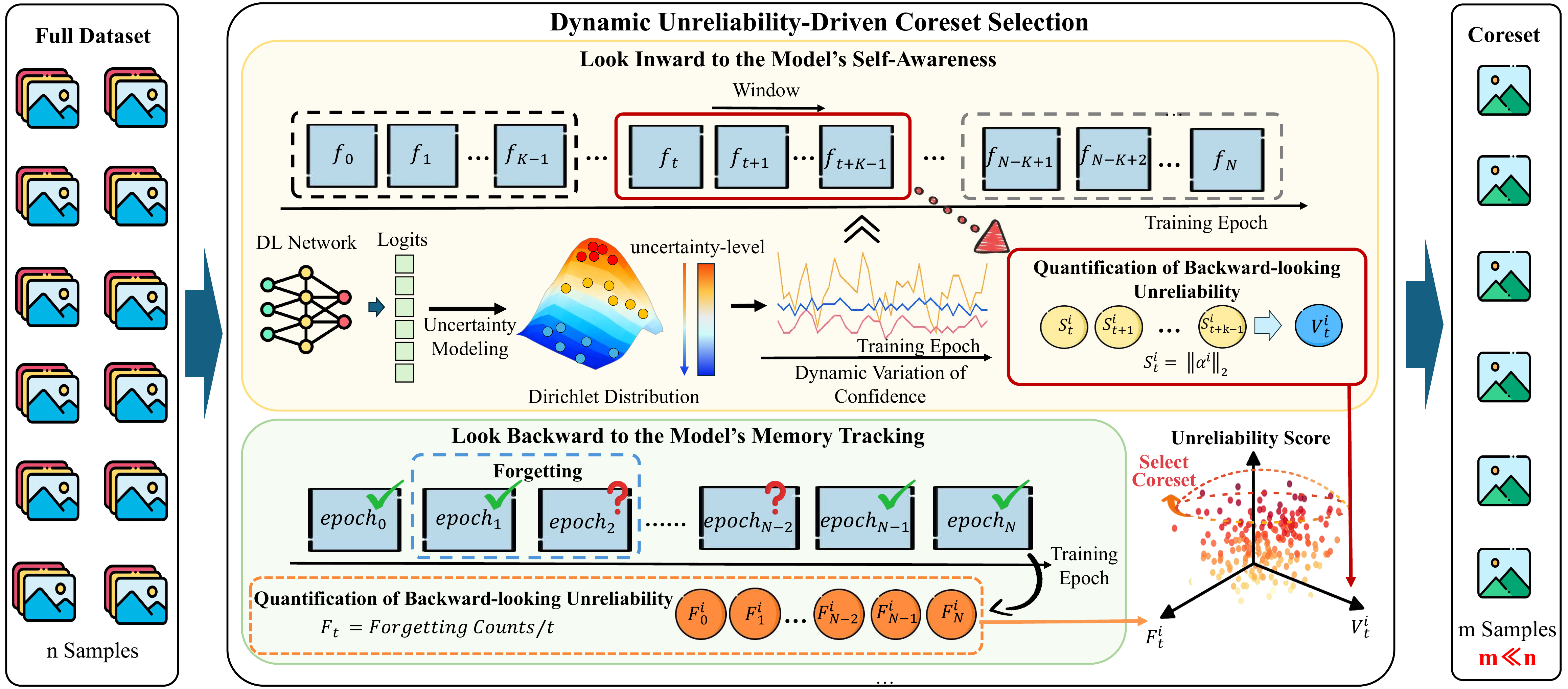}
    \vspace{-15pt}
    \caption{The framework of our proposed \textbf{DUCS}. \textbf{DUCS} is built upon a novel inward-backward unreliability assessment perspective. (1) To \textbf{look inward at the model’s self-awareness}, we parameterize the final layer outputs as concentration parameters of a Dirichlet distribution to quantify the epistemic uncertainty. This allows us to compute an epistemic uncertainty score \(S_{t}^{i}\) for each sample during training and subsequently calculate its variance \(V_t^{i}\) over a sliding window of epochs. (2) To \textbf{look backward at the model's memory tracking}, we calculate the forgetting frequency \(F_t^{i}\) throughout the training history. Ultimately, the coreset is constructed by selecting samples that exhibit both high variance in epistemic uncertainty and high forgetting frequency.}
    \vspace{-15pt}
    \label{fig:framework}
\end{figure*}

\section{Method}
\subsection{Problem Statement}
\label{subsec: coreset}
Let \(\mathbb{T} = \{(\bm{x^i}, \bm{y^i})\}_{i=1}^N\) denote the full training set, where \(\bm{x^i}\in \mathbb{R}^D\) is the input feature vector with $D = H \times W \times C$ representing the flattened input dimensionality, and \(\bm{y^i}\in \mathbb{R}^{1 \times C}\) is the one-hot encoded ground-truth label for \(C\) classes. All samples are assumed to be independently and identically distributed, drawn from an underlying distribution \(\mathcal{P}\). We define a neural network model as \(f_\theta\), parameterized by the weight vector \(\bm{\theta}\). The goal of coreset selection is to construct a subset \(\mathbb{S} \subset \mathbb{T}\) of size \(M = \lceil \beta N \rceil\), where \(\beta \in (0, 1]\) is the selection rate, such that a model trained on \(\mathbb{S}\) achieves performance comparable to that trained on the full dataset \(\mathbb{T}\)~\cite{sener2017active}. Formally, we aim to satisfy:

\begin{equation}
\label{eq:coreset} 
\mathop{\mathbb{E}}_{\substack{\theta_{0}\sim\mathcal{G} \\ (\bm{x},\bm{y})\sim \mathcal{P}}} \left[\ell\left(\bm{x},\bm{y};f_{\bm{\theta_{0}}}^{\mathbb{S}}\right)\right] \approx \mathop{\mathbb{E}}_{\substack{\bm{\theta_{0}}\sim\mathcal{G} \\ (\bm{x},\bm{y})\sim \mathcal{P}}} \left[\ell\left(\bm{x},\bm{y};f_{\bm{\theta_{0}}}^{\mathbb{T}}\right)\right],
\end{equation}
where \(f_{\bm{\theta_{0}}}^{\mathbb{S}}\) and \(f_{\bm{\theta_{0}}}^{\mathbb{T}}\) present models trained on \(\mathbb{S}\) and \(\mathbb{T}\), respectively, with initial parameters \(\bm{\theta_{0}}\) drawn from a distribution \(\mathcal{G}\).

To select informative samples from the dataset, we introduce a novel method, and the overview is presented in Section~\ref{subsec:over}. Subsequently, Section~\ref{subsec: inward} and~\ref{subsec: baward} introduce the two core components of DUCS: the inward-looking perspective, which models confidence dynamics, and the backward-looking perspective, which quantifies memory tracking. Finally, Section~\ref{subsec: unreliable score} integrates these components to present the Unreliability Score, which serves as the criterion for sample selection.

\subsection{Overview}
\label{subsec:over}
This section details our proposed DUCS method, with an overview presented in Figure~\ref{fig:framework}. Our proposed DUCS attempts to select coreset samples from both the inward- and backward-looking perspectives. Specifically, the inward-looking perspective equips the model with self-awareness regarding its predictive confidence through modeling confidence as a Dirichlet distribution. This allows us to select samples with significant confidence fluctuations, typically near decision boundaries. For the backward-looking perspective, DUCS measures a sample's forgetting frequency by tracking the evolution of its prediction accuracy throughout training. Samples that are frequently forgotten indicate high memory unreliability, which implies their proximity to decision boundaries. Finally, DUCS integrates confidence variance and forgetting frequency into an \textbf{Unreliability Score}, which quantifies each sample’s unreliability. This score is then used for coreset selection, prioritizing samples that carry informative signals for modeling decision boundaries.

\subsection{Inward-Looking Self-Awareness}
\label{subsec: inward}
The model is expected to self-quantify its confidence when predicting different samples. To achieve this, we parameterize the neural network outputs as the parameters of a Dirichlet distribution $D(\bm{p^i}\mid\bm{\alpha^i})$, rather than as deterministic probabilities, yielding smoother decision boundaries and enhanced generalization. The Dirichlet distribution for $x^i$ with parameters $\boldsymbol{\alpha}^{i}=[\alpha_1^{i}, ..., \alpha_C^{i}]$ can be obtained using the logits generated by the network as:
\begin{equation}
\label{eq:alpha_i}
\bm{\alpha^{i}} = \text{ReLU}(\bm{z^{i}}) + \bm{1},
\end{equation}
where \(\bm{z^{i}} = f_{\bm{\theta_{0}}}(\bm{x^{i}})\) represents the logits for the $i$-th sample, activated by the ReLU function and incremented by 1 to satisfy the parameter requirements of the Dirichlet distribution. Additionally, $z_c^{i}$ denotes the parameter of the Dirichlet distribution corresponding to the $c$-th class of $x^i$.
The parameter $\alpha_c^{i}$ refers to the confidence that $x^i$ is classified into the $c$-th class.

$\boldsymbol{p}^{i} = [p_1^{i}, \dots, p_C^{i}]$ denotes the predicted probability distribution for $\bm{x^i}$. The predicted probability of the $c$-th class for $\bm{x^i}$ is modeled as the proportion of $\alpha_c^i$, defined as $p_c^i=\frac{\alpha_c^i}{Q^i}$, where \(Q^i=\sum_{c=1}^C \alpha_c^i\) represents the Dirichlet strength for $\bm{x^i}$, reflecting the total support provided by the model for that sample. The larger \(Q^i\), the more sufficient the model's support for $\bm{x^i}$. 

To push the predicted probabilities $\boldsymbol{p^i}$ of $\bm{x^i}$ close to the ground truth $\boldsymbol{y^i}$, we employ the cross-entropy loss combined with the Dirichlet distribution density function, formulated as:
\begin{equation}
\begin{split}
    \mathcal{L}_{dirichlet} &=\int\left[\sum_{c=1}^{C}-y_{c}^{i} \log \left(p_{c}^{i}\right)\right] D(\boldsymbol{p^{i}} \mid \boldsymbol{\alpha^{i}}) d \boldsymbol{p^{i}} \\
    &=\sum_{c=1}^{C} y_{c}^{i}\left(\psi\left({Q}^{i}\right)-\psi\left({\alpha}_{c}^{i}\right)\right),
\end{split}
\end{equation}
where
\begin{equation}
    D(\boldsymbol{p^{i}} \mid \boldsymbol{\alpha^{i}}) = 
    \{
    \begin{array}{ll}
        \frac{1}{B(\boldsymbol{\alpha^i})} \prod_{c=1}^C{(p_{c}^{i})}^{\alpha_c^i-1} & \text{for } \boldsymbol{p^{i}} \in \mathcal{S}_{C}, \\
        0 & \text{otherwise}.
    \end{array}
\end{equation}

$B(\boldsymbol{\alpha^i})$ represents the $C$-dimension multinomial beta function with parameter $\boldsymbol{\alpha^{i}}$ and $\mathcal{S}_{C}$ denotes the $C$-class simplex. $y_c^{i}$ stands for the label for the $c$-th class of $\bm{x^i}$. The digamma function $\psi(\cdot)$ transforms the Dirichlet parameters into values for the optimization loss. To reduce the support of negative samples, we employ the Kullback-Leibler~(KL) divergence to penalize the $\alpha$ of incorrect classes:
\begin{equation}
\begin{aligned}
    \mathcal{L}_{K L}
    &=K L\left[
    D\left(\boldsymbol{p^{i}}\mid \hat{\boldsymbol{\alpha}^{i}}\right)
    \|
    D\left(\boldsymbol{p^{i}} \mid \boldsymbol{1} \right)\right] 
    \\
    &=\log \left(\frac{\Gamma\left(\sum_{c=1}^{C} \hat{{\alpha}}_{c}^{i}\right)}{\Gamma(C) \sum_{c=1}^{C} \Gamma\left(\hat{\alpha}_{c}^{i}\right)}\right)\\
    &+\sum_{c=1}^{C}\left(\hat{{\alpha}}_{c}^{i}-1\right)\left[\psi\left(\hat{{\alpha}}_{c}^{i}\right)-\psi\left(\sum_{c=1}^{C} \hat{{\alpha}}_{c}^{i}\right)\right],
\end{aligned}
\end{equation}
where the Gamma function $\Gamma(\cdot)$ generalizes the factorial to real numbers and is used to compute the difference between two Dirichlet distributions in the above KL divergence. \(\bm{\hat{\alpha}^i}\) is the parameter obtained by removing the support signals for incorrect classes using the true label of the sample, calculated by Eq.~\eqref{eq:hat_alpha_i}:
\begin{equation}
\label{eq:hat_alpha_i}
\bm{\hat{\alpha}^i}=(\bm{\alpha^i}-\bm{1})\cdot(\bm{1}-\bm{y^i})+\bm{1}.
\end{equation}

Finally, the loss is defined to improve the model's classification accuracy and self-awareness evaluation capability:
\begin{equation}
\label{eq:compound loss}
\mathcal{L} = - \sum_{i=1}^N\sum_{c=1}^C y_{c}^ilog(z_{c}^i) + \mathcal{L}_{dirichlet} + \lambda_t {\mathcal{L}_{K L}},
\end{equation}
where \(\lambda_t\) is the annealing coefficient for the current training epoch.

Based on the above model training process, the model learns how to quantify its confidence. At the $t$-th epoch, we compute the confidence score \(S_{t}^{i}= ||\boldsymbol{\alpha^i} ||_2\) for $\bm{x^{i}}$. This process generates a series of confidence scores that reveal the model's epistemic state with respect to the sample across different training iterations. To assess the stability of this epistemic state, we calculate the variance of these confidence scores over a series of consecutive epochs. Specifically, a sliding window of size \(K\) is introduced to analyze the dynamics of the confidence sequence \(({S_{t}^{i}, S_{t+1}^{i}, ..., S_{t+K-1}^{i}})\) for $\bm{x^{i}}$ from the $t$-th epoch to the $(t+K-1)$-th epoch. The variance of this sequence, denoted as ${V}_{t}^{i}$, quantifies the confidence fluctuation: lower variance indicates stability, while higher variance reflects greater uncertainty. This can be formulated as:
\begin{equation}
\label{eq:V_t}
{V}_{t}^{i}=\frac{1}{K} \sum_{k=t}^{t+K-1}\left(S_{k}^{i}-\frac{1}{K} \sum_{k=t}^{t+K-1} S_{k}^{i}\right)^{2}.
\end{equation}

\subsection{Backward-Looking Memory Tracking}

\label{subsec: baward}
In addition to examining the network's prediction confidence, we further explore its memory tracking to quantify the unreliability of samples from a backward-looking perspective.
Specifically, for the \(t\)-th training epoch on the full dataset, we record the prediction correctness \(d_t^{i} \in \{0,1\}\) for $\bm{x^i}$, where \(d_t^{i}=1\) indicates a correct prediction at the \(t\)-th epoch, and \(d_t^{i}=0\) indicates an incorrect prediction. A forgetting event for $\bm{x^i}$ occurs at the \(t\)-th epoch if and only if it was correctly classified at the \((t-1)\)-th epoch but incorrectly classified at the \(t\)-th epoch. This can be formulated as follows:
\begin{equation}
\label{eq:gamma_t}  
\gamma_t^{i} = 
\begin{cases} 
1, & \text{if } d_{t-1}^{i}=1 \text{ and } d_t^{i}=0, \\
0, & \text{otherwise}.
\end{cases}
\end{equation}

Subsequently, the cumulative number of forgetting events for $\bm{x^i}$ up to the \(t\)-th epoch is given by:
\begin{equation}
\label{eq:R_t}  
R_t^{i} = \sum_{z=1}^t \gamma_z^{i},
\end{equation}
where \(z\) is the summation index. The forgetting frequency of $\bm{x^i}$ at the $t$-th epoch, denoted as $F_t^{i}$, is defined as:
\begin{equation}
\label{eq:F_t}  
F_t^{i} =  \frac{R_t^{i}}{t}.
\end{equation}

Notably, if $\bm{x^i}$ has never been correctly classified in any training epoch up to and including the \(t\)-th epoch, then its forgetting frequency \(F_t^{i} = 1\). The forgetting frequency focuses on the dynamic behavior of how the model forgets samples during training, rather than the static cumulative count of forgetting events. This effectively reflects the model's memory tracking, where a high forgetting frequency indicates high memory unreliability for that sample, suggesting that the model's prediction for the sample is unstable during training or the sample lies near the decision boundary.

\begin{wrapfigure}{r}{0.66\textwidth}
  \centering
  \vspace{-20pt}
  
  \begin{subfigure}[b]{0.32\textwidth}
    \centering
    \includegraphics[width=\textwidth]{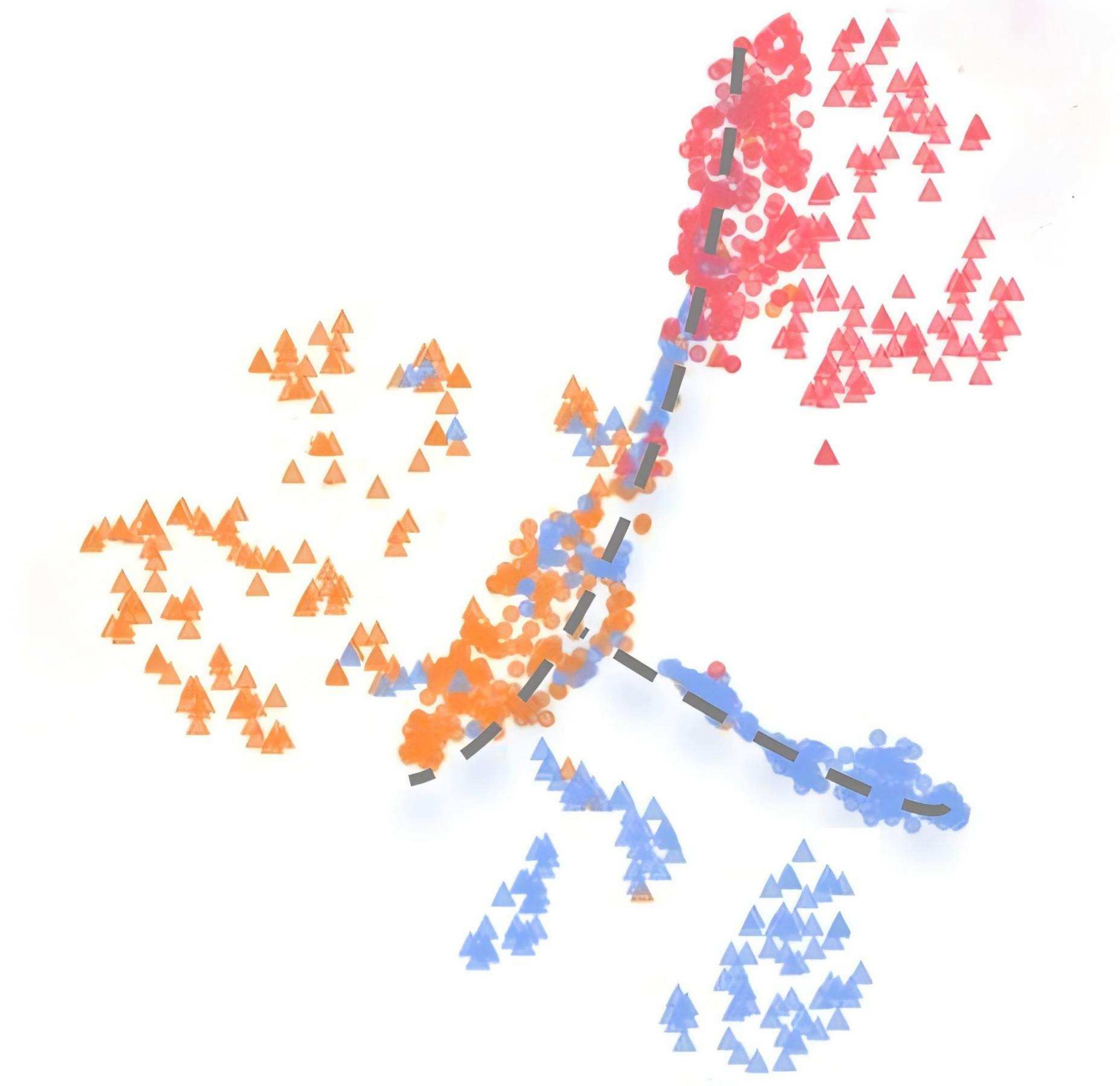}
    \caption{{\tiny Inward-Looking Perspective}}
    \label{fig:smnist-t-sne-s}
  \end{subfigure}
  \begin{subfigure}[b]{0.32\textwidth}
    \centering
    \includegraphics[width=\textwidth]{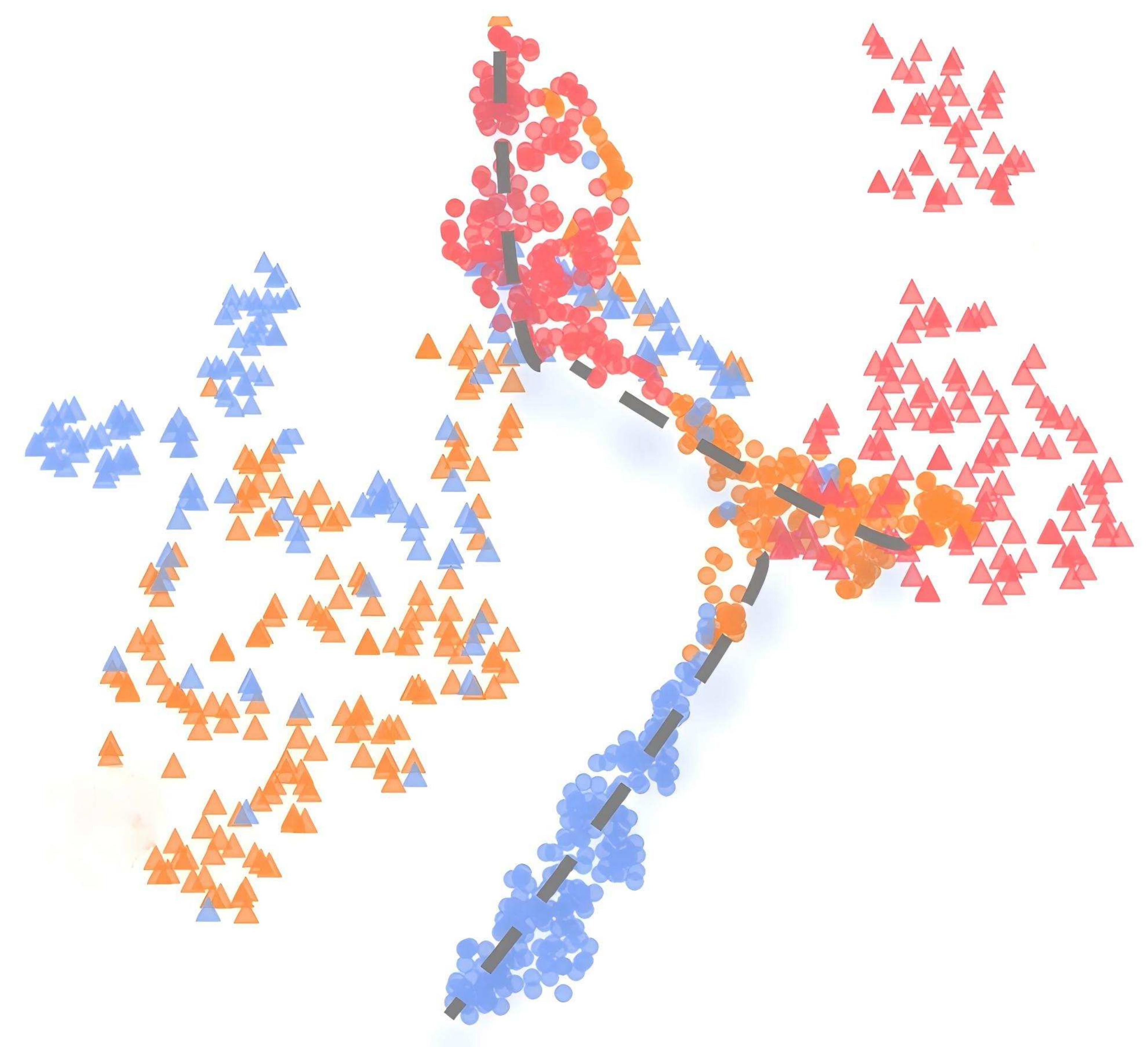}
    \caption{{\tiny Backward-Looking Perspective}} 
    \label{fig:smnist-t-sne-f}
  \end{subfigure}

  \vspace{-5pt}
  \caption{t-SNE visualizations. ``$\CIRCLE$'' and ``$\blacktriangle$'' denote top 20\% unreliable and reliable samples.}
  \label{fig:t-sne}
  \vspace{-20pt} 
\end{wrapfigure}

Figure.~\ref{fig:t-sne} provides a straightforward illustration of our motivation. It can be observed that samples with highly fluctuating confidence or high forgetting frequency tend to lie near the decision boundaries, whereas more reliable samples remain closer to the class centers. Therefore, DUCS attempts to jointly consider both perspectives in selecting a coreset, thereby facilitating the construction of decision boundaries.

\subsection{Unreliability Score}
\label{subsec: unreliable score}
Building upon two complementary perspectives, model self-awareness and memory tracking, we propose a unified evaluation metric, termed the \textit{\textbf{Unreliability Score}}. This metric integrates the epistemic uncertainty variance \(V_t^{i}\), derived from model introspection, and the forgetting frequency \(F_t^{i}\), which reflects the model's memory tracking. By combining these two factors, the unreliability score provides a comprehensive quantification of sample informativeness for coreset selection.

As analyzed above, samples exhibiting high confidence fluctuation or frequent forgetting are typically located near decision boundaries. 
Such samples are highly valuable for constructing robust decision boundaries.
Specifically, we focus on the forgetting frequency at the conclusion of training to quantify the backward-looking unreliability level. In this way, we can assess the unreliability level in the entire training process
This enables us to assess the forgetting dynamics across training. Once the model approaches convergence, with its parameters and predictions largely stabilized, samples that continue to be forgotten indicate persistent sensitivity to model updates and are likely to lie near the decision boundaries. This approach effectively mitigates the interference of transient fluctuations during training, ensuring that the constructed coreset most effectively supports the formation of accurate decision boundaries.

Based on the above analysis, we define the Unreliability Score \(U^{i} \) as:
\begin{equation}
\label{eq:U}
U^{i} = wF^{i}+V_t^{i},
\end{equation}
where $w$ is the temperature coefficient used to balance the two metrics and align their scales.

We then rank all samples in \(\mathbb{T}\) based on their Unreliability Score \(U^{i}\). Following~\cite{hong2024evolution}, we perform a grid search to determine the optimal window selection for different datasets and selection ratios. Samples with higher scores are considered more effective in supporting the formation of decision boundaries. Given a selection rate \(\beta\), we select the top-ranked \(M\) samples to form the coreset, where \(M = \lceil \beta N \rceil\).

\section{Experiments}
\label{sec:formatting}

This section presents comprehensive experiments to validate the effectiveness of our proposed DUCS. DUCS is first compared with various SOTA methods to evaluate its generalization and performance across multiple datasets in Section~\ref{subsec: baseline}. Subsequently, ablation studies are conducted to assess the contributions of different components in Section~\ref{subsec: ablation}.

\begin{table*}[!t]
\centering
\small
\caption{The Performance of ResNet-18 using various coreset selection methods on the MedMNIST medical datasets. All training runs were repeated three times with different random seeds to calculate the mean accuracy along with the standard deviation. The first and second results in each column are highlighted in red and blue, respectively.}
\vspace{-5pt}
\label{tab:benchmark_evaluation_results}
\resizebox{\textwidth}{!}{
\begin{tabular}{c|ccccccc|ccccccc}
\toprule
& \multicolumn{7}{c}{\textbf{OrganAMNIST}} & \multicolumn{7}{|c}{\textbf{OrganSMNIST}} \\
$\beta$ & 0.5\% & 1\% & 2\% & 5\% & 10\% & 20\% & 30\% & 0.5\% & 1\% & 2\% & 5\% & 10\% & 20\% & 30\% \\
\midrule
Full dataset& \multicolumn{7}{c}{98.39{\color{mygray}\footnotesize$\pm$0.02}} & \multicolumn{7}{|c}{91.76{\color{mygray}\footnotesize$\pm$0.55}} \\
\midrule
Random & 75.57 & 84.68 & 87.63 & 93.43 & 95.68 & 97.30 & 98.14 & 48.48 & 53.27 & 58.74 & 73.10 & 80.95 & 85.77 & 87.64 \\
 & {\color{mygray}\footnotesize$\pm$2.26} & {\color{mygray}\footnotesize$\pm$2.75} & {\color{mygray}\footnotesize$\pm$0.76} & {\color{mygray}\footnotesize$\pm$0.65} & {\color{mygray}\footnotesize$\pm$0.45} & {\color{mygray}\footnotesize$\pm$0.13} & {\color{mygray}\footnotesize$\pm$0.13} & {\color{mygray}\footnotesize$\pm$5.6} & {\color{mygray}\footnotesize$\pm$3.74} & {\color{mygray}\footnotesize$\pm$0.76} & {\color{mygray}\footnotesize$\pm$1.84} & {\color{mygray}\footnotesize$\pm$0.66} & {\color{mygray}\footnotesize$\pm$1.14} & {\color{mygray}\footnotesize$\pm$0.72} \\
Forgetting~\cite{toneva2018empirical} & 10.32 & 14.38 & 15.58 & 38.53 & 75.85 & 97.22 & 98.11 & 4.35 & 4.12 & 4.33 & 22.33 & 33.15 & 64.43 & 81.28 \\
 & {\color{mygray}\footnotesize$\pm$0.53} & {\color{mygray}\footnotesize$\pm$0.63} & {\color{mygray}\footnotesize$\pm$0.47} & {\color{mygray}\footnotesize$\pm$2.78} & {\color{mygray}\footnotesize$\pm$1.69} & {\color{mygray}\footnotesize$\pm$0.38} & {\color{mygray}\footnotesize$\pm$0.04} & {\color{mygray}\footnotesize$\pm$0.02} & {\color{mygray}\footnotesize$\pm$0.21} & {\color{mygray}\footnotesize$\pm$0.22} & {\color{mygray}\footnotesize$\pm$0.31} & {\color{mygray}\footnotesize$\pm$0.60} & {\color{mygray}\footnotesize$\pm$1.23} & {\color{mygray}\footnotesize$\pm$2.31} \\
Entropy~\cite{coleman2019selection} & 11.88 & 16.70 & 41.46 & 55.37 & 69.04 & 77.07 & 91.98 & 23.28 & 30.57 & 27.93 & 41.69 & 59.86 & 78.69 & 86.20 \\
 & {\color{mygray}\footnotesize$\pm$1.09} & {\color{mygray}\footnotesize$\pm$0.48} & {\color{mygray}\footnotesize$\pm$3.46} & {\color{mygray}\footnotesize$\pm$1.40} & {\color{mygray}\footnotesize$\pm$1.16} & {\color{mygray}\footnotesize$\pm$1.29} & {\color{mygray}\footnotesize$\pm$0.83} & {\color{mygray}\footnotesize$\pm$5.86} & {\color{mygray}\footnotesize$\pm$2.60} & {\color{mygray}\footnotesize$\pm$2.08} & {\color{mygray}\footnotesize$\pm$0.73} & {\color{mygray}\footnotesize$\pm$1.84} & {\color{mygray}\footnotesize$\pm$2.13} & {\color{mygray}\footnotesize$\pm$0.54} \\
EL2N~\cite{paul2021deep} & 7.74 & 9.25 & 14.16 & 40.68 & 81.25 & 97.25 & 98.16 & 7.55 & 12.6 & 17.63 & 23.24 & 28.24 & 37.58 & 60.06 \\
 & {\color{mygray}\footnotesize$\pm$1.02} & {\color{mygray}\footnotesize$\pm$0.45} & {\color{mygray}\footnotesize$\pm$1.14} & {\color{mygray}\footnotesize$\pm$3.36} & {\color{mygray}\footnotesize$\pm$3.22} & {\color{mygray}\footnotesize$\pm$0.24} & {\color{mygray}\footnotesize$\pm$0.30} & {\color{mygray}\footnotesize$\pm$1.12} & {\color{mygray}\footnotesize$\pm$4.84} & {\color{mygray}\footnotesize$\pm$1.59} & {\color{mygray}\footnotesize$\pm$1.88} & {\color{mygray}\footnotesize$\pm$1.44} & {\color{mygray}\footnotesize$\pm$1.53} & {\color{mygray}\footnotesize$\pm$2.14} \\
AUM~\cite{pleiss2020identifying} & 10.01 & 10.30 & 12.81 & 35.10 & 68.44 & 93.76 & 98.12 & 5.48 & 4.16 & 4.56 & 7.01 & 22.13 & 39.87 & 65.93 \\
 & {\color{mygray}\footnotesize$\pm$0.41} & {\color{mygray}\footnotesize$\pm$1.73} & {\color{mygray}\footnotesize$\pm$2.62} & {\color{mygray}\footnotesize$\pm$3.46} & {\color{mygray}\footnotesize$\pm$0.95} & {\color{mygray}\footnotesize$\pm$1.89} & {\color{mygray}\footnotesize$\pm$0.14} & {\color{mygray}\footnotesize$\pm$2.14} & {\color{mygray}\footnotesize$\pm$0.14} & {\color{mygray}\footnotesize$\pm$0.18} & {\color{mygray}\footnotesize$\pm$1.24} & {\color{mygray}\footnotesize$\pm$1.86} & {\color{mygray}\footnotesize$\pm$2.19} & {\color{mygray}\footnotesize$\pm$1.61} \\
CCS~\cite{zheng2022coverage} & \textcolor{blue}{86.34} & \textcolor{blue}{87.79} & 88.05 & 93.51 & 95.58 & 96.86 & 97.18 & \textcolor{blue}{45.46} & \textcolor{blue}{55.48} & 58.43 & 71.73 & 78.46 & 83.64 & 84.94 \\
 & {\color{mygray}\footnotesize$\pm$0.72} & {\color{mygray}\footnotesize$\pm$0.60} & {\color{mygray}\footnotesize$\pm$0.62} & {\color{mygray}\footnotesize$\pm$0.10} & {\color{mygray}\footnotesize$\pm$0.32} & {\color{mygray}\footnotesize$\pm$0.25} & {\color{mygray}\footnotesize$\pm$0.08} & {\color{mygray}\footnotesize$\pm$4.79} & {\color{mygray}\footnotesize$\pm$5.45} & {\color{mygray}\footnotesize$\pm$0.25} & {\color{mygray}\footnotesize$\pm$0.83} & {\color{mygray}\footnotesize$\pm$0.18} & {\color{mygray}\footnotesize$\pm$0.55} & {\color{mygray}\footnotesize$\pm$0.22} \\
Moderate~\cite{xia2023moderate} & 40.74 & 45.31 & 54.42 & 75.38 & 86.47 & 91.66 & 92.54 & 22.40 & 32.20 & 35.41 & 52.51 & 59.73 & 67.90 & 73.05 \\
 & {\color{mygray}\footnotesize$\pm$1.36} & {\color{mygray}\footnotesize$\pm$1.26} & {\color{mygray}\footnotesize$\pm$0.86} & {\color{mygray}\footnotesize$\pm$1.06} & {\color{mygray}\footnotesize$\pm$1.30} & {\color{mygray}\footnotesize$\pm$1.08} & {\color{mygray}\footnotesize$\pm$0.93} & {\color{mygray}\footnotesize$\pm$1.22} & {\color{mygray}\footnotesize$\pm$0.94} & {\color{mygray}\footnotesize$\pm$1.32} & {\color{mygray}\footnotesize$\pm$1.62} & {\color{mygray}\footnotesize$\pm$1.53} & {\color{mygray}\footnotesize$\pm$1.62} & {\color{mygray}\footnotesize$\pm$1.53} \\
TDDS~\cite{zhang2024spanning} & 27.93 & 31.93 & 59.72 & 82.49 & 90.51 & 94.53 & 94.96 & 26.08 & 27.62 & 44.77 & 55.69 & 62.90 & 68.60 & 74.27 \\
 & {\color{mygray}\footnotesize$\pm$0.20} & {\color{mygray}\footnotesize$\pm$0.98} & {\color{mygray}\footnotesize$\pm$1.01} & {\color{mygray}\footnotesize$\pm$0.77} & {\color{mygray}\footnotesize$\pm$1.33} & {\color{mygray}\footnotesize$\pm$1.94} & {\color{mygray}\footnotesize$\pm$1.61} & {\color{mygray}\footnotesize$\pm$3.26} & {\color{mygray}\footnotesize$\pm$1.71} & {\color{mygray}\footnotesize$\pm$1.98} & {\color{mygray}\footnotesize$\pm$1.16} & {\color{mygray}\footnotesize$\pm$1.60} & {\color{mygray}\footnotesize$\pm$0.62} & {\color{mygray}\footnotesize$\pm$1.09} \\
EVA~\cite{hong2024evolution} & 76.14 & 84.93 & \textcolor{blue}{88.83} & \textcolor{blue}{94.43} & \textcolor{blue}{97.20} & \textcolor{blue}{98.27} & \textcolor{blue}{98.63} & 42.57 & 54.1 & \textcolor{blue}{61.23} & \textcolor{blue}{78.71} & \textcolor{blue}{83.11} & \textcolor{blue}{86.38} & \textcolor{blue}{88.77} \\
 & {\color{mygray}\footnotesize$\pm$3.32} & {\color{mygray}\footnotesize$\pm$2.22} & {\color{mygray}\footnotesize$\pm$0.88} & {\color{mygray}\footnotesize$\pm$1.32} & {\color{mygray}\footnotesize$\pm$0.34} & {\color{mygray}\footnotesize$\pm$0.57} & {\color{mygray}\footnotesize$\pm$0.34} & {\color{mygray}\footnotesize$\pm$2.45} & {\color{mygray}\footnotesize$\pm$3.94} & {\color{mygray}\footnotesize$\pm$0.75} & {\color{mygray}\footnotesize$\pm$0.93} & {\color{mygray}\footnotesize$\pm$0.72} & {\color{mygray}\footnotesize$\pm$1.02} & {\color{mygray}\footnotesize$\pm$0.43} \\
\midrule
\rowcolor{lightgray}
\textbf{DUCS (Ours)} & \textcolor{red!90!black}{87.38} & \textcolor{red!90!black}{90.99} & \textcolor{red!90!black}{93.27} & \textcolor{red!90!black}{96.50} & \textcolor{red!90!black}{97.66} & \textcolor{red!90!black}{98.44} & \textcolor{red!90!black}{98.68} & \textcolor{red!90!black}{58.18} & \textcolor{red!90!black}{72.45} & \textcolor{red!90!black}{77.17} & \textcolor{red!90!black}{84.47} & \textcolor{red!90!black}{86.71} & \textcolor{red!90!black}{89.45} & \textcolor{red!90!black}{89.90} \\
\rowcolor{lightgray}
 & {\color{mygray}\footnotesize$\pm$0.36} & {\color{mygray}\footnotesize$\pm$0.11} & {\color{mygray}\footnotesize$\pm$0.08} & {\color{mygray}\footnotesize$\pm$0.08} & {\color{mygray}\footnotesize$\pm$0.06} & {\color{mygray}\footnotesize$\pm$0.04} & {\color{mygray}\footnotesize$\pm$0.09} & {\color{mygray}\footnotesize$\pm$0.11} & {\color{mygray}\footnotesize$\pm$0.49} & {\color{mygray}\footnotesize$\pm$0.16} & {\color{mygray}\footnotesize$\pm$0.19} & {\color{mygray}\footnotesize$\pm$0.18} & {\color{mygray}\footnotesize$\pm$1.55} & {\color{mygray}\footnotesize$\pm$0.11} \\
\bottomrule
\end{tabular}
}
\vspace{-10pt}
\end{table*}

\begin{table}[t] 
\centering
\small

\begin{minipage}{0.46\textwidth}
    \centering
    \caption{High selection rate performance on OrganAMNIST and OrganSMNIST with ResNet-18. The first and second best results in each column are marked in red and blue, respectively. }
    \label{tab: High Ratio}
    \vspace{-6pt}
    \resizebox{\linewidth}{!}{
        {\setlength{\tabcolsep}{2pt}
        
 \begin{tabular}{c|ccc|ccc}
    \toprule
    & \multicolumn{3}{c}{\textbf{OrganAMNIST}} & \multicolumn{3}{|c}{\textbf{OrganSMNIST}} \\
    $\beta$ & 50\% & 70\% & 90\% & 50\% & 70\% & 90\% \\
    \midrule
    Full dataset& \multicolumn{3}{c}{98.39{\color{mygray}\tiny$\pm$0.02}} & \multicolumn{3}{|c}{91.76{\color{mygray}\tiny$\pm$0.55}} \\
    \midrule
Random & 98.14 & 98.29 & 98.44 & 89.63 & 90.23 & 91.16 \\
 & {\color{mygray}\footnotesize$\pm$0.04} & {\color{mygray}\footnotesize$\pm$0.15} & {\color{mygray}\footnotesize$\pm$0.43} & {\color{mygray}\footnotesize$\pm$0.68} & {\color{mygray}\footnotesize$\pm$0.72} & {\color{mygray}\footnotesize$\pm$0.23} \\
Forgetting~\cite{toneva2018empirical} & 98.46 & 98.31 & \textcolor{blue}{98.70} & 91.18 & 91.46 & 91.58 \\
 & {\color{mygray}\footnotesize$\pm$0.16} & {\color{mygray}\footnotesize$\pm$0.82} & {\color{mygray}\footnotesize$\pm$0.44} & {\color{mygray}\footnotesize$\pm$0.69} & {\color{mygray}\footnotesize$\pm$0.38} & {\color{mygray}\footnotesize$\pm$0.26} \\
Entropy~\cite{coleman2019selection} & 97.93 & 98.32 & 98.50 & 90.41 & 91.31 & 91.36 \\
 & {\color{mygray}\footnotesize$\pm$0.44} & {\color{mygray}\footnotesize$\pm$0.03} & {\color{mygray}\footnotesize$\pm$0.55} & {\color{mygray}\footnotesize$\pm$0.27} & {\color{mygray}\footnotesize$\pm$0.50} & {\color{mygray}\footnotesize$\pm$0.78} \\
EL2N~\cite{paul2021deep} & 98.39 & 98.24 & 98.48 & 89.94 & 90.43 & 91.53\\
 & {\color{mygray}\footnotesize$\pm$0.89} & {\color{mygray}\footnotesize$\pm$0.04} & {\color{mygray}\footnotesize$\pm$1.27} & {\color{mygray}\footnotesize$\pm$0.57} & {\color{mygray}\footnotesize$\pm$0.90} & {\color{mygray}\footnotesize$\pm$0.61} \\
AUM~\cite{pleiss2020identifying} & 98.13 & 98.54 & 98.55 & 89.81 & 90.97 & 91.58 \\
 & {\color{mygray}\footnotesize$\pm$0.02} & {\color{mygray}\footnotesize$\pm$0.76} & {\color{mygray}\footnotesize$\pm$0.51} & {\color{mygray}\footnotesize$\pm$0.52} & {\color{mygray}\footnotesize$\pm$0.87} & {\color{mygray}\footnotesize$\pm$0.03} \\
CCS~\cite{zheng2022coverage} & 97.55 & 97.56 & 97.07 & 85.53 & 85.69 & 86.38 \\
 & {\color{mygray}\footnotesize$\pm$0.19} & {\color{mygray}\footnotesize$\pm$0.60} & {\color{mygray}\footnotesize$\pm$0.81} & {\color{mygray}\footnotesize$\pm$0.11} & {\color{mygray}\footnotesize$\pm$0.15} & {\color{mygray}\footnotesize$\pm$0.91} \\
EVA~\cite{hong2024evolution} & \textcolor{blue}{98.80} & \textcolor{blue}{98.96} & 98.68 & \textcolor{blue}{91.21} & \textcolor{blue}{91.94} & \textcolor{blue}{91.89} \\
 & {\color{mygray}\footnotesize$\pm$0.56} & {\color{mygray}\footnotesize$\pm$0.73} & {\color{mygray}\footnotesize$\pm$1.45} & {\color{mygray}\footnotesize$\pm$0.98} & {\color{mygray}\footnotesize$\pm$0.74} & {\color{mygray}\footnotesize$\pm$0.96} \\
Moderate~\cite{xia2023moderate} & 93.48 & 94.11 & 95.08 & 75.64 & 80.60 & 81.70 \\
 & {\color{mygray}\footnotesize$\pm$0.53} & {\color{mygray}\footnotesize$\pm$0.58} & {\color{mygray}\footnotesize$\pm$0.52} & {\color{mygray}\footnotesize$\pm$0.78} & {\color{mygray}\footnotesize$\pm$2.28} & {\color{mygray}\footnotesize$\pm$1.90} \\
TDDS~\cite{zhang2024spanning} & 95.05 & 95.41 & 95.55 & 75.07 & 75.51 & 76.02 \\
 & {\color{mygray}\footnotesize$\pm$1.44} & {\color{mygray}\footnotesize$\pm$1.37} & {\color{mygray}\footnotesize$\pm$1.67} & {\color{mygray}\footnotesize$\pm$1.32} & {\color{mygray}\footnotesize$\pm$1.81} & {\color{mygray}\footnotesize$\pm$1.47} \\
 
\midrule
\rowcolor{lightgray}
\textbf{DUCS (Ours)} & \textcolor{red!90!black}{98.91} & \textcolor{red!90!black}{99.10} & \textcolor{red!90!black}{99.13} & \textcolor{red!90!black}{91.82} & \textcolor{red!90!black}{92.18} & \textcolor{red!90!black}{92.65} \\
\rowcolor{lightgray}
 & {\color{mygray}\footnotesize$\pm$0.11} & {\color{mygray}\footnotesize$\pm$0.13} & {\color{mygray}\footnotesize$\pm$0.14} & {\color{mygray}\footnotesize$\pm$0.16} & {\color{mygray}\footnotesize$\pm$0.33} & {\color{mygray}\footnotesize$\pm$0.17} \\
    \bottomrule
    \end{tabular}
    }
}
\end{minipage}
\hfill
\begin{minipage}{0.52\textwidth}
    \centering
    \caption{The performance on natural image datasets CIFAR-10 and CIFAR-100 with ResNet-18. The first and second best results in each column are highlighted in red and blue, respectively.}
    \label{tab:cifar}
    \vspace{-5pt}
    \renewcommand{\arraystretch}{1.03}
    \resizebox{\linewidth}{!}{
        {\setlength{\tabcolsep}{1.2pt}
        \begin{tabular}{c|cccc|cccc}
        \toprule
        & \multicolumn{4}{c}{\textbf{CIFAR-10}} & \multicolumn{4}{|c}{\textbf{CIFAR-100}} \\
        $\beta$ & 2\% & 5\% & 10\% & 20\% & 2\% & 5\% & 10\% & 20\% \\
        \midrule
Full dataset & \multicolumn{4}{c}{92.24{\color{mygray}\tiny$\pm$0.74}} & \multicolumn{4}{|c}{78.19{\color{mygray}\tiny$\pm$0.23}} \\
\midrule
Random & 41.64 & 58.62 & 71.64 & 85.57 & 13.35 & 20.53 & 37.10 & 53.68\\
 & {\color{mygray}\footnotesize$\pm$0.92} & {\color{mygray}\footnotesize$\pm$0.29} & {\color{mygray}\footnotesize$\pm$0.47} & {\color{mygray}\footnotesize$\pm$0.33} & {\color{mygray}\footnotesize$\pm$0.39} & {\color{mygray}\footnotesize$\pm$0.93} & {\color{mygray}\footnotesize$\pm$1.01} & {\color{mygray}\footnotesize$\pm$1.33} \\
Forgetting~\cite{toneva2018empirical} & 36.20 & 41.67 & 52.29 & 76.00 & 6.86 & 10.14 & 16.87 & 26.18\\
 & {\color{mygray}\footnotesize$\pm$0.24} & {\color{mygray}\footnotesize$\pm$0.58} & {\color{mygray}\footnotesize$\pm$0.33} & {\color{mygray}\footnotesize$\pm$1.45} & {\color{mygray}\footnotesize$\pm$0.08} & {\color{mygray}\footnotesize$\pm$0.32} & {\color{mygray}\footnotesize$\pm$0.12} & {\color{mygray}\footnotesize$\pm$0.61} \\
Entropy~\cite{coleman2019selection} & 32.08 & 47.70 & 60.52 & 75.69 & 8.92 & 14.64 & 25.01  & 40.33 \\
 & {\color{mygray}\footnotesize$\pm$0.42} & {\color{mygray}\footnotesize$\pm$0.39} & {\color{mygray}\footnotesize$\pm$0.15} & {\color{mygray}\footnotesize$\pm$0.90} & {\color{mygray}\footnotesize$\pm$0.40} & {\color{mygray}\footnotesize$\pm$0.50} & {\color{mygray}\footnotesize$\pm$0.46} & {\color{mygray}\footnotesize$\pm$0.24} \\
EL2N~\cite{paul2021deep} & 10.54 & 15.94 & 23.45 & 42.62 & 3.63 & 5.16 & 7.26 & 14.65\\
 & {\color{mygray}\footnotesize$\pm$0.49} & {\color{mygray}\footnotesize$\pm$0.61} & {\color{mygray}\footnotesize$\pm$0.86} & {\color{mygray}\footnotesize$\pm$0.68} & {\color{mygray}\footnotesize$\pm$0.02} & {\color{mygray}\footnotesize$\pm$0.22} & {\color{mygray}\footnotesize$\pm$0.22} & {\color{mygray}\footnotesize$\pm$0.87} \\
AUM~\cite{pleiss2020identifying} & 14.66 & 18.54 & 25.35 & 49.51 & 3.92 & 5.25 & 8.38 & 16.64\\
 & {\color{mygray}\footnotesize$\pm$0.52} & {\color{mygray}\footnotesize$\pm$0.59} & {\color{mygray}\footnotesize$\pm$0.22} & {\color{mygray}\footnotesize$\pm$1.10} & {\color{mygray}\footnotesize$\pm$0.03} & {\color{mygray}\footnotesize$\pm$0.04} & {\color{mygray}\footnotesize$\pm$0.29} & {\color{mygray}\footnotesize$\pm$0.07} \\
CCS~\cite{zheng2022coverage} & 43.95 & 51.45 & 71.78 & 85.53 & 13.50 & 23.83 & 36.39 & 53.14\\
 & {\color{mygray}\footnotesize$\pm$1.66} & {\color{mygray}\footnotesize$\pm$2.18} & {\color{mygray}\footnotesize$\pm$1.98} & {\color{mygray}\footnotesize$\pm$0.97} & {\color{mygray}\footnotesize$\pm$0.47} & {\color{mygray}\footnotesize$\pm$1.07} & {\color{mygray}\footnotesize$\pm$1.94} & {\color{mygray}\footnotesize$\pm$1.34} \\
EVA~\cite{hong2024evolution} & \textcolor{blue}{46.27} & \textcolor{blue}{61.75} & 73.73 & 85.12 & 13.28 & 24.38 & 39.60 & 55.86 \\
 & {\color{mygray}\footnotesize$\pm$0.37} & {\color{mygray}\footnotesize$\pm$0.57} & {\color{mygray}\footnotesize$\pm$0.42} & {\color{mygray}\footnotesize$\pm$0.68} & {\color{mygray}\footnotesize$\pm$0.33} & {\color{mygray}\footnotesize$\pm$0.87} & {\color{mygray}\footnotesize$\pm$0.46} & {\color{mygray}\footnotesize$\pm$0.92} \\
Moderate~\cite{xia2023moderate} & 44.03 & 57.65 & 69.05 & 79.31 & 11.50 & 21.12 & 32.38 & 55.30\\
 & {\color{mygray}\footnotesize$\pm$0.88} & {\color{mygray}\footnotesize$\pm$0.79} & {\color{mygray}\footnotesize$\pm$1.78} & {\color{mygray}\footnotesize$\pm$0.86} & {\color{mygray}\footnotesize$\pm$0.60} & {\color{mygray}\footnotesize$\pm$0.82} & {\color{mygray}\footnotesize$\pm$1.72} & {\color{mygray}\footnotesize$\pm$0.86} \\
TDDS~\cite{zhang2024spanning} & 44.21 & 55.67 & \textcolor{blue}{76.52} & \textcolor{red!90!black}{88.61} & \textcolor{blue}{16.15} & \textcolor{blue}{29.26} & \textcolor{blue}{45.45} & \textcolor{red!90!black}{59.88} \\
 & {\color{mygray}\footnotesize$\pm$0.68} & {\color{mygray}\footnotesize$\pm$1.15} & {\color{mygray}\footnotesize$\pm$0.62} & {\color{mygray}\footnotesize$\pm$0.84} & {\color{mygray}\footnotesize$\pm$0.19} & {\color{mygray}\footnotesize$\pm$0.84} & {\color{mygray}\footnotesize$\pm$0.90} & {\color{mygray}\footnotesize$\pm$1.00} \\
\midrule
\rowcolor{lightgray}
\textbf{DUCS (Ours)} & \textcolor{red!90!black}{53.10} & \textcolor{red!90!black}{68.47} & \textcolor{red!90!black}{79.60} & \textcolor{blue}{86.64} & \textcolor{red!90!black}{18.84} & \textcolor{red!90!black}{32.22} & \textcolor{red!90!black}{46.23} & \textcolor{blue}{58.30}\\
\rowcolor{lightgray}
 & {\color{mygray}\footnotesize$\pm$0.15} & {\color{mygray}\footnotesize$\pm$1.04} & {\color{mygray}\footnotesize$\pm$1.68} & {\color{mygray}\footnotesize$\pm$0.76} & {\color{mygray}\footnotesize$\pm$0.75} & {\color{mygray}\footnotesize$\pm$1.08} & {\color{mygray}\footnotesize$\pm$0.63} & {\color{mygray}\footnotesize$\pm$0.56} \\
        \bottomrule
        \end{tabular}
        }
    }
\end{minipage}
\vspace{-10pt}
\end{table}

\subsection{Experiment Setup}
\textbf{Datasets.} MedMNIST~\cite{yang2023medmnist} is a large-scale medical image classification dataset, comprising 10 sub-datasets that cover multiple modalities, data scales, and classification tasks. Due to the high training cost, we focus on two 2D datasets from MedMNIST: OrganAMNIST and OrganSMNIST. Each dataset targets a multi-class classification task involving 11 organs: bladder, left/right femur, heart, left/right kidney, liver, left/right lung, pancreas, and spleen. 

\noindent \textbf{Baselines and Networks.} We compare our approach with nine representative baselines: Random Selection, Forgetting Score~\cite{toneva2018empirical}, Entropy~\cite{coleman2019selection}, EL2N~\cite{paul2021deep}, Area Under the Margin~(AUM)~\cite{pleiss2020identifying}, Coverage-Centric Coreset Selection~(CCS)~\cite{zheng2022coverage}, Evolution-aware VAriance~(EVA)~\cite{hong2024evolution}, Moderate~\cite{xia2023moderate} and Temporal Dual-depth Scoring~(TDDS)~\cite{zhang2024spanning}. Our main evaluations are conducted using ResNet-18~\cite{he2016deep}, while cross-architecture generalization is further verified with ResNet-50~\cite{he2016deep}, MobileNet~\cite{sandler2018mobilenetv2}, and LeNet~\cite{lecun2002gradient}.

\noindent \textbf{Implementation details.} To ensure fairness in our comparisons, we follow the experimental setup outlined in~\cite{hong2024evolution}. All experiments are implemented in PyTorch~\cite{paszke2017automatic} and conducted on a single NVIDIA RTX 3090 GPU. Unless stated otherwise in our paper, both the coreset and surrogate networks adopt ResNet-18 with identical hyperparameters before and after selection. The surrogate network is trained for 200 epochs on all datasets. For inward-looking unreliability quantification, the start and end epochs of each window are determined using a grid search with a step size of 10.

\subsection{Benchmark Evaluation Results}
\label{subsec: baseline}

\textbf{Performance Comparison and Analysis.} The results of the systematic comparison between DUCS and other SOTA methods on both OrganSMNIST and OrganAMNIST are presented in Table~\ref{tab:benchmark_evaluation_results}. It can be seen that our method yields the best performance in this comparison, especially under the challenging selection ratio scenario.
Besides, on OrganAMNIST, DUCS performs nearly identically to full-data training at a 10\% selection ratio, and even outperforms it at the 20\% and 30\% selection ratios. Additionally, under extremely low selection ratios (0.5\% and 1\%), DUCS continues to show clear advantages, exceeding the random baseline by 11.81\% and 6.31\% on OrganAMNIST, and by 9.70\% and 19.18\% on OrganSMNIST. These results confirm that DUCS effectively identifies and retains the most informative samples for model training, even in scenarios with highly limited data.

As mentioned earlier, medical data often exhibit high intra-class variance and strong inter-class similarity, which inherently blur decision boundaries. DUCS addresses this by selecting informative samples to construct accurate decision boundaries, as illustrated in Figure.~\ref{fig:t-sne}. In contrast, other methods struggle to maintain their performance under challenging selection ratios, mainly because they fail to sufficiently capture the intrinsic uncertainty of decision-boundary samples. As a result, these methods demonstrate a sampling bias toward overconfident or redundant regions, which ultimately weakens the model’s ability to preserve discriminative decision boundaries when working with limited data. Further evaluations on PathMNIST~\cite{yang2023medmnist} and the high-resolution CT dataset, COVID-LDCT~\cite{afshar2022human}, can be found in the \textbf{\textit{Appendix}}.

\noindent \textbf{Performance at High Selection Ratios and Small-Scale Dataset.} In this work, our experiments primarily focus on ultra-low and low selection ratios to evaluate the effectiveness of DUCS under data-scarce conditions. We also investigate the performance of the proposed method under high selection ratios, and the results are presented in Table~\ref{tab: High Ratio}. It can be seen that DUCS maintains competitive performance even at high selection ratios. 
Furthermore, the effectiveness of DUCS extends beyond medical imaging. Preliminary experiments on well-established natural image datasets on CIFAR-10 and CIFAR-100~\cite{krizhevsky2009learning} demonstrate that DUCS outperforms current SOTA methods, and the results are provided in Table~\ref{tab:cifar}.
In addition, we have further validated its adaptability and robustness on the smaller medical dataset PneumoniaMNIST, and the results are shown in Table~\ref{tab: PneumoniaMNIST}.

\begin{wraptable}{r}{0.5\textwidth} 
\centering
\vspace{-35pt} 
\caption{Performance on the small-size medical imaging dataset PneumoniaMNIST with ResNet-18.}
\label{tab: PneumoniaMNIST}

\resizebox{\linewidth}{!}{
    {\setlength{\tabcolsep}{6pt}
    \begin{tabular}{c|cccc}
    \toprule
    & \multicolumn{4}{c}{\textbf{PneumoniaMNIST}} \\
    $\beta$ & 10\% & 30\% & 50\% & 70\% \\
    \midrule
    Full dataset & \multicolumn{4}{c}{97.27} \\
    \midrule
    Random & 92.77 & 95.70 & 95.90 & 96.48 \\
    Forgetting~\cite{toneva2018empirical} & 55.86 & 96.48 & 95.70 & 95.52 \\
    Entropy~\cite{coleman2019selection} & 71.48 & 96.09 & 95.70 & 97.07 \\
    EL2N~\cite{paul2021deep} & 26.17 & 94.34 & 97.07 & 95.90 \\
    AUM~\cite{pleiss2020identifying} & 38.87 & 96.29 & 96.29 & 96.68 \\
    CCS~\cite{zheng2022coverage} & 91.02 & 91.21 & 92.97 & 91.02 \\
    Moderate~\cite{xia2023moderate} & 88.14 & 90.82 & 94.23 & 95.33 \\
    TDDS~\cite{zhang2024spanning} & 82.85 & 90.02 & 91.67 & 94.18 \\
    EVA~\cite{hong2024evolution} & \textcolor{red!90!black}{94.34} & \textcolor{blue}{97.07} & \textcolor{blue}{97.46} & \textcolor{blue}{97.66} \\
    \midrule
    \rowcolor{lightgray}
    \textbf{DUCS (Ours)} & \textcolor{blue}{93.32} & \textcolor{red!90!black}{97.33} & \textcolor{red!90!black}{97.71} & \textcolor{red!90!black}{98.28} \\
    \bottomrule
    \end{tabular}
    }
}
\vspace{-15pt} 
\end{wraptable}

\noindent \textbf{Generalization across Architectures.} The previous experiments assumed that the coreset and surrogate networks share the same architecture. In this section, we relax this assumption and consider the case where these networks have different architectures. Specifically, a ResNet-18 model is first trained on the full dataset, and multiple scoring strategies are applied to select coresets under different selection ratios. 
These coresets are subsequently used to train three representative architectures: ResNet-50, MobileNet-v2, and LeNet. The results, shown in Tables~\ref{tab: Cross-architecture in OrganSMNIT} and~\ref{tab: Cross-architecture in OrganAMNIT}, show that DUCS outperforms most SOTA approaches and demonstrates strong transferability across diverse architectures, particularly under ultra-low selection ratios (0.5\%, 1\%, and 2\%). Moreover, our method consistently exhibits robust and superior generalization performance even under high selection ratios. Additionally, we conduct experiments with different network architectures, such as Transformer-based models. The corresponding results are reported in the \textbf{Appendix}.

\begin{table*}[ht]
\centering
\small
\vspace{-20pt}
\caption{Cross-architecture generalization performance. We train ResNet-50, LeNet, and MobileNet-v2 models with coresets of OrganSMNIST selected by scores calculated on training dynamics with ResNet-18.}
\vspace{-5pt}
\label{tab: Cross-architecture in OrganSMNIT}
\resizebox{\textwidth}{!}{%
\setlength{\tabcolsep}{2.5pt}
\begin{tabular}{@{}c|cccccc|cccccc|cccccc@{}}
\toprule
& \multicolumn{6}{c}{\textbf{ResNet-50}} & \multicolumn{6}{c}{\textbf{LeNet}} & \multicolumn{6}{c}{\textbf{MobileNet-v2}} \\
$\beta$ & 0.5\% & 1\% & 2\% & 5\% & 10\% & 20\% & 0.5\% & 1\% & 2\% & 5\% & 10\% & 20\% & 0.5\% & 1\% & 2\% & 5\% & 10\% & 20\% \\
\midrule
Forgetting~\cite{toneva2018empirical} & 4.49 & 4.24 & 4.05 & 4.64 & 35.79 & 59.42 & 7.67 & 7.67 & 3.87 & 4.59 & 24.76 & 33.59 & 4.24 & 4.24 & 4.08 & 4.10 & 28.66 & 53.47 \\
Entropy~\cite{coleman2019selection} & 20.39 & 22.02 & 24.22 & 26.71 & 48.29 & 76.90 & 16.27 & 20.71 & 21.83 & 21.58 & 47.56 & 65.92 & 20.68 & 22.50 & 23.54 & 24.80 & 54.15 & 73.00 \\
EL2N~\cite{paul2021deep} & 7.32 & 6.40 & 13.70 & 15.33 & 56.64 & 67.24 & 79.98 & 11.91 & 15.74 & 14.60 & 46.92 & 61.08 & 18.56 & 16.97 & 15.99 & 20.56 & 57.99 & 52.34 \\
AUM~\cite{pleiss2020identifying} & 4.24 & 4.24 & 4.24 & 4.10 & 22.36 & 37.16 & 3.87 & 4.00 & 4.28 & 4.30 & 13.38 & 32.08 & 4.24 & 4.69 & 4.77 & 4.00 & 21.44 & 37.55 \\
CCS~\cite{zheng2022coverage} & 31.97 & \textcolor{blue}{47.68} & \textcolor{blue}{44.86} & 43.90 & 71.88 & 78.27 & \textcolor{blue}{40.01} & 39.80 & \textcolor{blue}{51.75} & 52.15 & 69.48 & 71.44 & \textcolor{blue}{47.39} & \textcolor{blue}{51.92} & \textcolor{blue}{58.03} & 45.65 & 77.98 & 81.40 \\
Moderate~\cite{xia2023moderate} & 23.23 & 26.47 & 37.22 & 51.14 & 57.47 & 66.20 & 17.56 & 23.85 & 25.21 & 41.29 & 49.31 & 54.14 & 23.88 & 31.70 & 35.50 & 49.64 & 60.42 & 69.60  \\
TDDS~\cite{zhang2024spanning} & 21.04 & 23.54 & 24.77 & 46.52 & 55.69 & 66.61 & 16.42 & 26.01 & 37.51 & 45.85 & 50.93 & 58.60  & 24.48 & 32.22 & 43.28 & 50.32 & 60.46 & 70.52  \\
EVA~\cite{hong2024evolution} & \textcolor{blue}{41.06} & 32.86 & 41.46 & \textcolor{blue}{52.82} & \textcolor{blue}{77.00} & \textcolor{blue}{82.96} & 28.17 & \textcolor{blue}{46.48} & 45.12 & \textcolor{blue}{59.67} & \textcolor{blue}{70.90} & \textcolor{red!90!black}{77.00} & 43.31 & 50.78 & 44.19 & \textcolor{blue}{51.17} & \textcolor{blue}{79.79} & \textcolor{blue}{84.38} \\
\midrule
\rowcolor{lightgray}
\textbf{DUCS (Ours)} & \textcolor{red!90!black}{51.39} & \textcolor{red!90!black}{61.46} & \textcolor{red!90!black}{64.85} & \textcolor{red!90!black}{76.95} & \textcolor{red!90!black}{81.08} & \textcolor{red!90!black}{87.03} & \textcolor{red!90!black}{44.41} & \textcolor{red!90!black}{46.70} & \textcolor{red!90!black}{62.52} & \textcolor{red!90!black}{63.13} & \textcolor{red!90!black}{71.90} & \textcolor{blue}{75.65} & \textcolor{red!90!black}{50.73} & \textcolor{red!90!black}{66.39} & \textcolor{red!90!black}{70.51} & \textcolor{red!90!black}{76.63} & \textcolor{red!90!black}{84.18} & \textcolor{red!90!black}{88.74} \\
\bottomrule
\end{tabular}
}
\vspace{-15pt}
\end{table*}

\begin{table*}[t]
\centering
\small
\caption{Cross-architecture generalization performance. We train ResNet-50, LeNet, and MobileNet-v2 models with coresets of OrganAMNIST selected by scores calculated on training dynamics with ResNet-18.}
\label{tab: Cross-architecture in OrganAMNIT}
\resizebox{\textwidth}{!}{%
\setlength{\tabcolsep}{2.5pt}
\begin{tabular}{@{}c|cccccc|cccccc|cccccc@{}}
\toprule
& \multicolumn{6}{c}{\textbf{ResNet-50}} & \multicolumn{6}{c}{\textbf{LeNet}} & \multicolumn{6}{c}{\textbf{MobileNet-v2}} \\
$\beta$ & 0.5\% & 1\% & 2\% & 5\% & 10\% & 20\% & 0.5\% & 1\% & 2\% & 5\% & 10\% & 20\% & 0.5\% & 1\% & 2\% & 5\% & 10\% & 20\% \\
\midrule
Forgetting~\cite{toneva2018empirical} & 11.00 & 12.91 & 12.42 & 34.52 & 68.33 & 96.09 & 8.75 & 12.09 & 13.61 & 23.93 & 47.33 & 83.79 & 8.79 & 11.89 & 14.14 & 25.08 & 71.25 & 95.59 \\
Entropy~\cite{coleman2019selection} & 9.77 & 16.78 & 15.96 & 19.06 & 34.71 & 72.93 & 4.70 & 13.28 & 10.69 & 17.45 & 26.84 & 40.24 & 9.84 & 15.65 & 18.87 & 19.26 & 37.73 & 69.00 \\
EL2N~\cite{paul2021deep} & 8.70 & 10.15 & 11.82 & 21.68 & 56.63 & 94.58 & 8.75 & 9.21 & 12.83 & 19.11 & 45.65 & 84.79 & 9.88 & 8.83 & 13.71 & 24.05 & 57.99 & 93.59 \\
AUM~\cite{pleiss2020identifying} & 13.25 & 9.97 & 11.86 & 30.60 & 34.96 & 86.52 & 9.81 & 8.75 & 9.51 & 20.60 & 32.72 & 61.93 & 9.43 & 11.52 & 12.51 & 20.18 & 33.15 & 89.91 \\
CCS~\cite{zheng2022coverage} & \textcolor{blue}{79.76} & \textcolor{blue}{87.95} & \textcolor{blue}{82.53} & \textcolor{blue}{92.33} & \textcolor{blue}{95.13} & \textcolor{blue}{96.70} & 29.75 & \textcolor{blue}{52.04} & 62.73 & \textcolor{blue}{80.43} & 75.87 & 80.76 & \textcolor{blue}{83.65} & \textcolor{blue}{88.21} & \textcolor{blue}{83.64} & \textcolor{blue}{94.04} & \textcolor{blue}{95.12} & \textcolor{blue}{97.60} \\
Moderate~\cite{xia2023moderate} & 38.75 & 43.68 & 51.59 & 70.18 & 83.88 & 89.22 & 33.74 & 34.20 & 38.40 & 52.29 & 67.13 & 76.73 & 43.02 & 48.45 & 57.67 & 75.02 & 86.06 & 91.48  \\
TDDS~\cite{zhang2024spanning} & 15.80 & 19.24 & 46.17 & 71.79 & 72.63 & 81.88 & 19.64 & 25.91 & 34.11 & 62.48 & 64.74 & 70.63 & 20.49 & 27.52 & 51.01 & 80.44 & 82.20 & 91.83  \\
EVA~\cite{hong2024evolution} & 76.04 & 78.48 & 76.16 & 89.89 & 95.05 & 96.03 & \textcolor{blue}{34.94} & 46.89 & \textcolor{blue}{68.48} & 73.03 & \textcolor{blue}{85.68} & \textcolor{red!90!black}{93.57} & 75.96 & 84.88 & 78.82 & 90.95 & 95.05 & 97.18 \\
\midrule
\rowcolor{lightgray}
\textbf{DUCS (Ours)} & \textcolor{red!90!black}{82.59} & \textcolor{red!90!black}{88.20} & \textcolor{red!90!black}{88.17} & \textcolor{red!90!black}{93.43} & \textcolor{red!90!black}{96.36} & \textcolor{red!90!black}{97.90} & \textcolor{red!90!black}{35.94} & \textcolor{red!90!black}{55.80} & \textcolor{red!90!black}{72.53} & \textcolor{red!90!black}{85.41} & \textcolor{red!90!black}{86.68} & \textcolor{blue}{90.80} & \textcolor{red!90!black}{84.53} & \textcolor{red!90!black}{89.46} & \textcolor{red!90!black}{88.72} & \textcolor{red!90!black}{95.02} & \textcolor{red!90!black}{97.21} & \textcolor{red!90!black}{98.06} \\
\bottomrule
\end{tabular}
}
\vspace{-5pt}
\end{table*}

\noindent \textbf{Evaluation on 3D Datasets.}  While previous experiments were performed on 2D medical imaging datasets, we further extend our evaluation to 3D medical imaging to validate the generalizability of DUCS. Specifically, three public 3D benchmarks from OrganMNIST3D, NoduleMNIST3D, and FractureMNIST3D datasets~\cite{yang2023medmnist} are selected to validate our proposed method.
They cover tasks ranging from 11-class organ classification to binary malignancy detection and 3-class fracture typing.

\begin{wraptable}{r}{0.45\textwidth}
    \centering
    \vspace{-10pt}
    \caption{Performance on 3D datasets across different selection ratios.}
    \label{tab:3d}
    \resizebox{\linewidth}{!}{
        {\setlength{\tabcolsep}{3pt}
        \begin{tabular}{c|cccc|c}
        \toprule
        $\beta$ & 5\% & 10\% & 20\% & 30\% & Full\\
        \midrule
        Organ3D  & 87.50 & 95.62 & 97.50 & 98.12 & 98.44\\
        Nodule3D & 78.12 & 81.88 & 76.88 & 83.75 & 82.81 \\
        Fracture3D & 48.00 & 48.75 & 51.25 & 52.50 & 53.91\\
        \bottomrule
        \end{tabular}
        }
    }
    \vspace{-20pt}
\end{wraptable}
For the 3D experiments, we adopt 3D ResNet-18 (R3D-18)~\cite{tran2018closer} as both the coreset and surrogate networks. As shown in Table~\ref{tab:3d}, DUCS achieves stable and impressive performance across different coreset ratios. On OrganMNIST3D, using only 5\% of the data already yields 87.5\% accuracy, and performance rapidly approaches the full-data accuracy of 98.44\% as the coreset grows. On NoduleMNIST3D, DUCS performs consistently across different ratios, with the best accuracy of 83.75\% at a 30\% coreset ratio. For the more challenging FractureMNIST3D dataset, DUCS shows steady improvement with larger coresets, ultimately approaching the full-data baseline. These results demonstrate that DUCS remains effective and robust for 3D medical imaging tasks.

\subsection{Ablation Study and Analysis}
\label{subsec: ablation}

\begin{wrapfigure}{r}{0.5\textwidth} 
\vspace{-25pt}
  \centering
  % \vspace{-10pt}
  \begin{subfigure}[b]{0.49\linewidth}
    \centering
    \includegraphics[width=\linewidth]{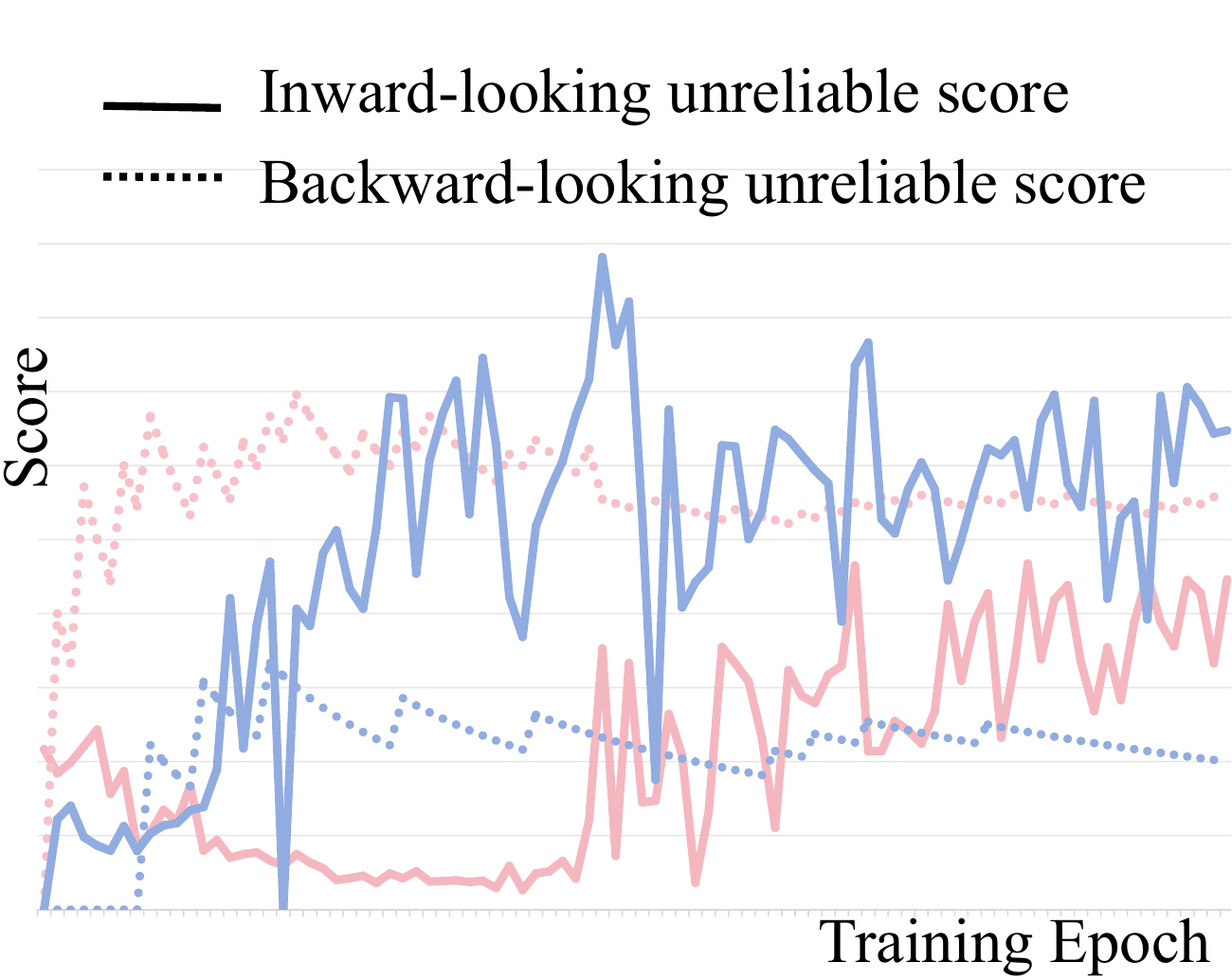}
    \caption{OrganSMNIST}
    \label{fig:score-s}
  \end{subfigure}
  \hfill
  \begin{subfigure}[b]{0.49\linewidth}
    \centering
    \includegraphics[width=\linewidth]{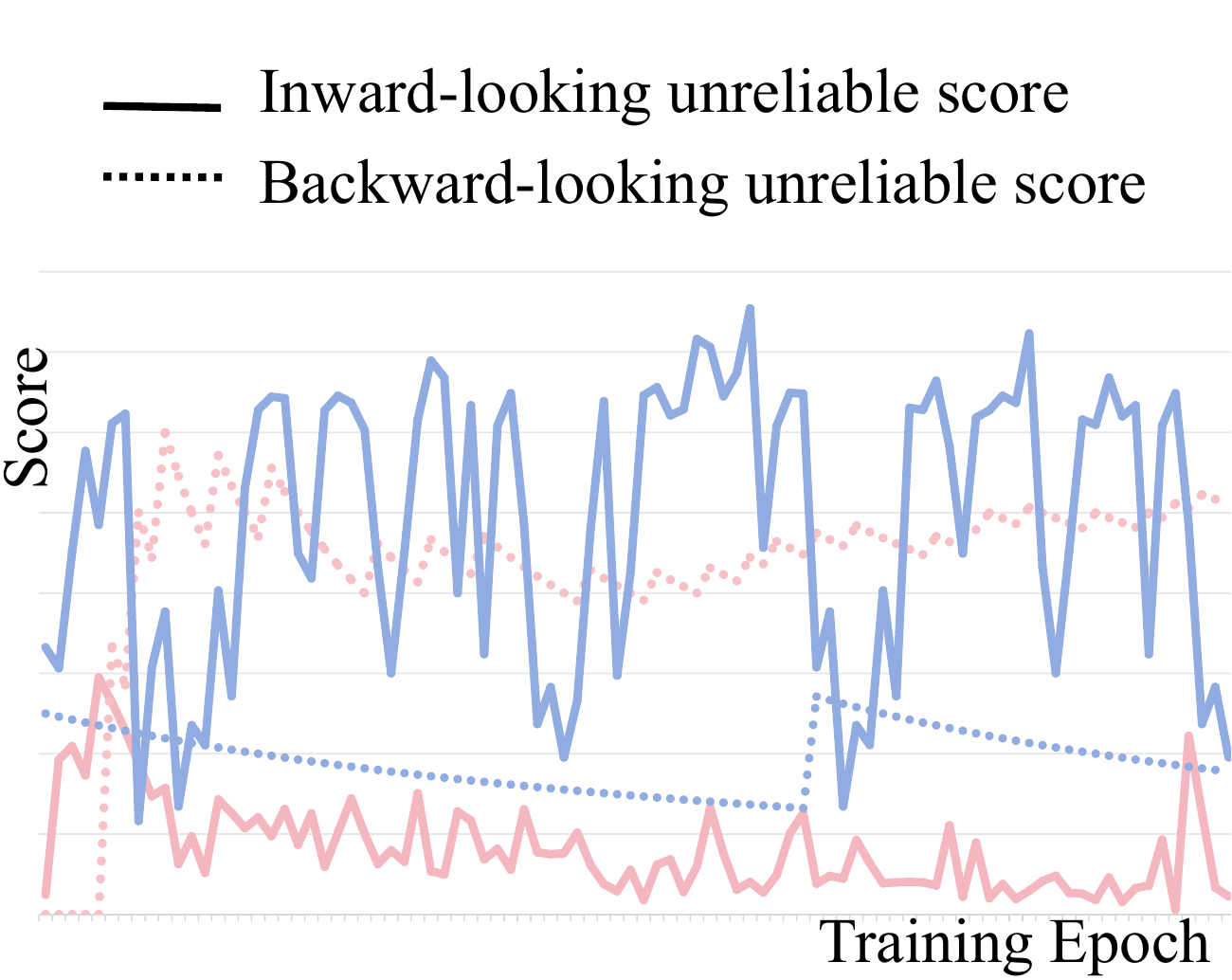}
    \caption{OrganAMNIST}
    \label{fig:score-a}
  \end{subfigure}
  \vspace{-15pt}
  \caption{The score variations of two randomly selected samples from OrganSMNIST and OrganAMNIST during the training process. Different colors represent different samples.}
  \label{fig:Score-Dynamics}
  \vspace{-20pt} 
\end{wrapfigure}

\noindent \textbf{Complementarity of Perspectives.} As illustrated in Figure~\ref{fig:Score-Dynamics}, 
we evaluate the  dynamic evolution of inward-looking unreliability score \(S_t^i\) and backward-looking unreliability score \(F_t^{i}\) during training. The results reveal that samples with high inward-looking scores do not always exhibit high forgetting frequency, and vice versa. This observation suggests that assessing sample importance from a single perspective is insufficient, as the two perspectives provide complementary insights into model behavior and sample reliability.

In our work, we select coreset samples by jointly considering both inward- and backward-looking perspectives. To assess their effectiveness, we first introduce a baseline that uses only inward-looking unreliability scores.
We further design ``DUCS$\dagger$", which selects coresets based on inward-looking unreliability scores but trains surrogate networks with a backward-looking–driven sampling strategy
\begin{wrapfigure}{l}{0.5\textwidth}
  \vspace{-10pt}
  \centering
  \begin{subfigure}[b]{0.49\linewidth}
    \centering
    \includegraphics[width=\linewidth]{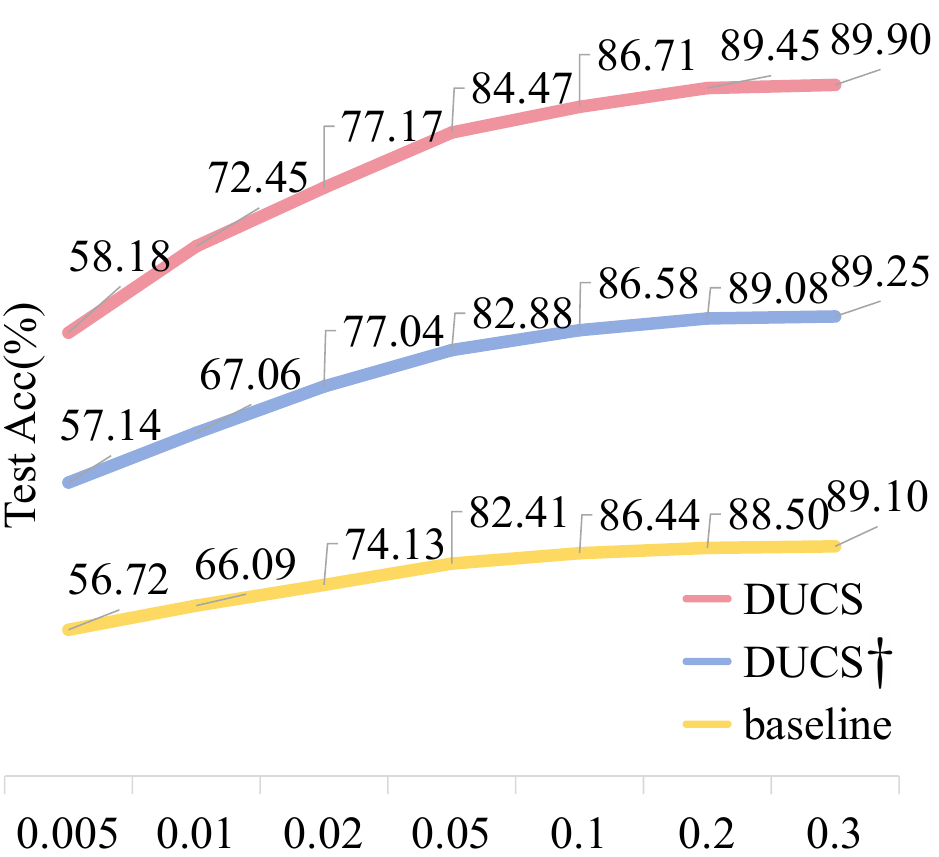}
    \caption{OrganSMNIST}
    \label{fig:ablation-s}
  \end{subfigure}
  \hfill
  \begin{subfigure}[b]{0.49\linewidth}
    \centering
    \includegraphics[width=\linewidth]{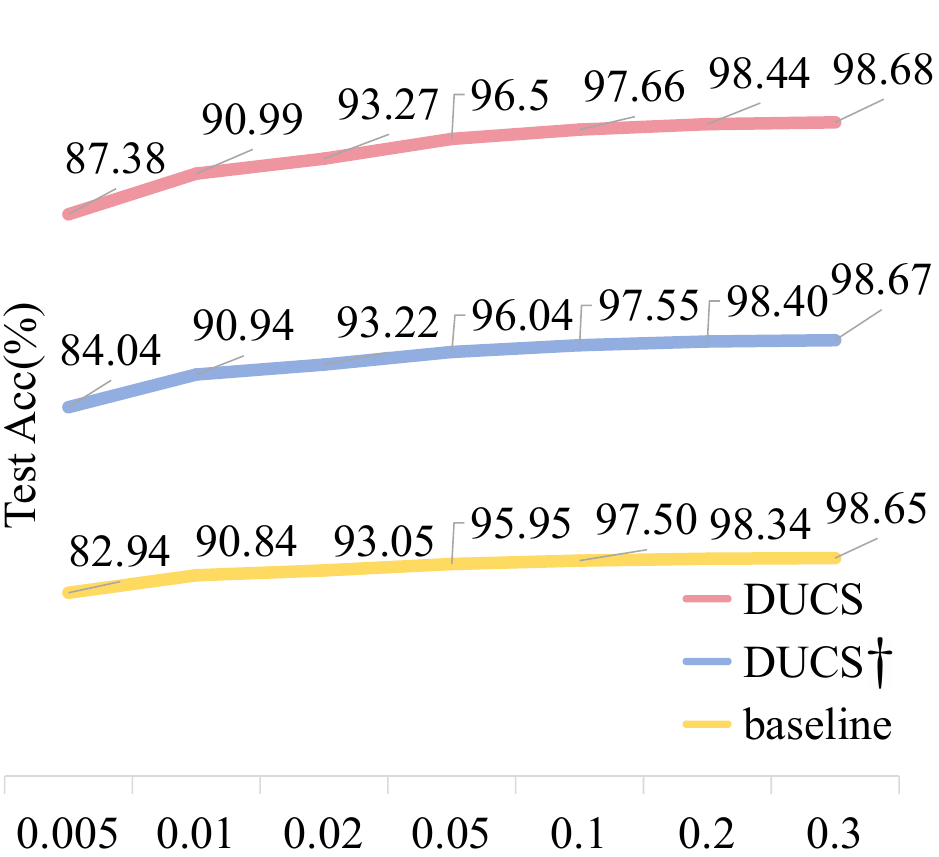}
    \caption{OrganAMNIST}
    \label{fig:ablation-a}
  \end{subfigure}
  \vspace{-15pt}
  \caption{We validate the effectiveness of evaluating sample importance from the two complementary perspectives on the OrganSMNIST and OrganAMNIST datasets.}
  \label{fig:ablation}
  \vspace{-20pt}
\end{wrapfigure}
that gives higher weight to more unreliable samples. As shown in Figure~\ref{fig:ablation}, performance consistently increases as the selection ratio grows, which aligns with intuition. Besides, incorporating the backward-looking perspective yields significant gains, confirming the benefit of combining both perspectives for reliable sample estimation.

\noindent \textbf{Window Size}
To assess the effect of the sliding window size $K$ on the performance, we conduct an ablation study on the OrganAMNIST and OrganSMNIST datasets, where $K \in\{5,10,20,40\}$. As shown in Table~\ref{tab: window_size}, $K=10$ consistently achieves the best overall performance across different coreset ratios $\beta$.

On the OrganAMNIST dataset, performance improves when increasing the
\begin{wraptable}{l}{0.75\textwidth} 
    \centering
    \vspace{-20pt}
    \caption{Ablation study on window size $K$ on the OrganAMNIST and OrganSMNIST. The best results in each column are in red.}
    \label{tab: window_size}
    \resizebox{\linewidth}{!}{
        {\setlength{\tabcolsep}{2pt} 
        \begin{tabular}{c|ccccccc|ccccccc}
        \toprule
         & \multicolumn{7}{c}{\textbf{OrganAMNIST}} & \multicolumn{7}{|c}{\textbf{OrganSMNIST}} \\
        $\beta$  & 0.5\% & 1\% & 2\% & 5\% & 10\% & 20\% & 30\% & 0.5\% & 1\% & 2\% & 5\% & 10\% & 20\% & 30\%  \\
        \midrule
        K=5 & \textcolor{red!90!black}{87.47} & \textcolor{red!90!black}{93.10} & \textcolor{red!90!black}{93.27} & 96.47        
        & \textcolor{red!90!black}{97.97}        
        & 98.29 &98.43 & \textcolor{red!90!black}{59.26} 
        & 72.38 & 75.08 & 83.03 & 86.70 & 89.07 &\textcolor{red!90!black}{89.91} \\
        K=10 &87.38 &90.99 & \textcolor{red!90!black}{93.27} & \textcolor{red!90!black}{96.50} & 97.66 & \textcolor{red!90!black}{98.44} & \textcolor{red!90!black}{98.68} &58.28 &\textcolor{red!90!black}{72.45} & \textcolor{red!90!black}{77.17} & \textcolor{red!90!black}{84.47} & \textcolor{red!90!black}{86.71} 
        & \textcolor{red!90!black}{89.45} &89.90 \\
        K=20 &85.25 &85.87 & 89.99 & 95.18 & 96.89 & 98.09 &98.66 &47.68 &61.95 & 70.96 & 80.46 & 85.03 & 88.78 &89.97 \\
        K=40 &79.25 &83.65 & 90.03 & 94.22 & 95.86 & 97.66 &98.68 &50.16 &64.36 & 69.37 & 80.75 & 83.61 & 88.46 &89.66 \\
        \bottomrule
        \end{tabular}
        }
    }
    \vspace{-20pt}
\end{wraptable}
window size from $K=5$ to $K=10$, while further enlarging $K$ to 20 or 40 leads to a performance drop. A similar pattern is observed on OrganSMNIST, where $K=10$ also achieves the best overall performance. These results suggest that extremely small window sizes may introduce fluctuations due to insufficient contextual information, whereas excessively large ones may dilute informative confidence cues. Therefore, $K=10$ provides a reasonable trade-off between capturing local variations and maintaining stable confidence estimation, and is adopted as the default window size in our paper.

\subsection{Analysis of the Computational Overhead}
To assess the practicality of DUCS in real-world applications, we evaluate its computational efficiency in this section. The computational costs of DUCS and other methods are summarized in Table~\ref{tab: Computational costs}, which compares the computational time required for score computation on an NVIDIA RTX 3090 GPU. Following~\cite{hong2024evolution}, we adopt EL2N~\cite{paul2021deep} as our baseline reference due to its high efficiency. As shown in the table, DUCS introduces minimal additional overhead compared to other baseline methods. Given its notable performance improvement, the added computational cost is well justified and remains acceptable.

Moreover, we separately evaluate the computational overhead of different components on the OrganSMNIST dataset, and compare the accuracy of individual components versus the full method across various data selection ratios. The corresponding results are reported in Table~\ref{tab:ap-computational-costs}. The backward-looking component is computationally efficient, requiring less than 1 second and introducing negligible additional overhead. The inward-looking component incurs slightly higher computational cost due to its confidence modeling process, but the overall runtime remains moderate. A detailed analysis of the computational overhead is provided in the \textbf{\textit{Appendix}}.

\begin{table}[!t]
    \centering
    \begin{minipage}[t]{0.47\textwidth}
        \centering
        \caption{\small Computational costs on calculating scores.}
        \vspace{-8pt}
        \label{tab: Computational costs}
        \renewcommand{\arraystretch}{0.99}
        \resizebox{\linewidth}{!}{
            {\setlength{\tabcolsep}{3pt} 
            \begin{tabular}{c|cc|cc}
            \toprule
            & \multicolumn{2}{c}{\textbf{OrganAMNIST}} & \multicolumn{2}{c}{\textbf{OrganSMNIST}}\\
              & Time (s) & $\Delta$EL2N & Time (s) & $\Delta$EL2N \\
            \midrule
            EL2N~\cite{paul2021deep} & 12.26 & - & 6.26 & - \\
            Forgetting~\cite{toneva2018empirical} & 15.10 & +23.2\% & 7.50 & +19.8\% \\
            Entropy~\cite{coleman2019selection} & 12.69 & +3.5\% & 8.14 & +30.0\% \\
            AUM~\cite{pleiss2020identifying} & 16.76 & +36.7\% & 8.28 & +32.3\% \\
            CCS~\cite{zheng2022coverage} & 16.78 & +36.9\% & 8.29 & +32.4\% \\
            Moderate~\cite{xia2023moderate} & 4.37 & -64.35\% & 2.59 & -58.62\% \\
            TDDS~\cite{zhang2024spanning} & 1.76 & -85.64\% & 0.89 & -85.78\% \\
            EVA~\cite{hong2024evolution} & 13.04 & +6.4\% & 6.52 & +4.2\% \\
            \midrule
            \rowcolor{lightgray}
            \textbf{DUCS (Ours)} & 15.99 & +30.4\% & 7.37 & +17.7\% \\
            \bottomrule
            \end{tabular}
            }
        }
    \end{minipage}
    \vspace{-10pt}
    \hfill 
    \begin{minipage}[t]{0.49\textwidth} 
        \centering
        \caption{\small Computational costs and classification accuracy across different components of DUCS under varying selection ratios ($\beta$) on the OrganSMNIST dataset.}
        \label{tab:ap-computational-costs}
        \renewcommand{\arraystretch}{1.2}
        \resizebox{\linewidth}{!}{
            {\setlength{\tabcolsep}{7pt}
            \begin{tabular}{c|c|c|c}
            \toprule
            & Backward-looking & Inward-looking & DUCS \\
            \midrule
            Time (s) & 0.73 & 6.64 & 7.37 \\
            \midrule
            $\beta=0.02$ & 47.72 & 74.13 & 77.17  \\
            $\beta=0.05$ & 63.21 & 82.41 & 84.47  \\
            $\beta=0.1$ & 73.41 & 86.44 & 86.71  \\
            $\beta=0.2$ & 85.64 & 88.50 & 89.45 \\
            \bottomrule
            \end{tabular}
            }
        }
    \end{minipage}
    \vspace{-10pt}
\end{table}

\section{Discussion}
\label{sec: discussion}
While the proposed DUCS strategy exhibits strong generalization and consistent performance across diverse datasets and architectures, yet there remain avenues for further research in the future. 
Although the computational overhead of DUCS is minimal relative to full training, recording dynamic statistics across epochs may limit scalability. Developing lightweight approximations or online update mechanisms would be valuable for improving efficiency without compromising selection accuracy.

\section{Conclusion}
\label{sec: conclusion}
We have identified a critical limitation in existing coreset selection for medical imaging: when confronted with substantial intra-class variations and high inter-class similarity, conventional strategies tend to over-select ``easy" samples located near class centers. This leads to model bias and blurred decision boundaries. To address this bottleneck, we propose an innovative strategy, DUCS. Our approach shifts the paradigm by strategically mining samples that the model deems ``unreliable", specifically those exhibiting both strong confidence fluctuations and high forgetting frequency. By synergistically integrating the inward- and backward-looking perspectives, DUCS achieves precise identification of critical samples near decision boundaries. Consequently, DUCS facilitates the construction of models with superior generalization capabilities, using only minimal data subsets. Extensive evaluations across various datasets and architectures demonstrate the superior performance of the proposed DUCS strategy.

\bibliographystyle{splncs04}
\bibliography{main}

\appendix
\clearpage
\setcounter{page}{1}
\setcounter{table}{0}
\renewcommand{\thetable}{S\arabic{table}}  
\setcounter{equation}{0}
\renewcommand{\theequation}{S\arabic{equation}} 
\setcounter{figure}{0}
\renewcommand{\thefigure}{S\arabic{figure}} 
\section{Algorithmic}

\begin{wrapfigure}{r}{0.5\linewidth}
\footnotesize
\vspace{-20pt} 
\rule{\linewidth}{0.8pt} 
\captionof{algorithm}{The main steps of the proposed DUCS.} 
\label{alg: DUCS}
\vspace{-5pt}
\rule{\linewidth}{0.4pt} 
\begin{algorithmic}[1]
\Statex \hspace{-1.2em}\textbf{Input:} Full training set $\mathbb{T}=\{(x^i,y^i)\}_{i=1}^N$, selection ratio $\beta$, window size $K$, training epochs $T$.
\Statex \hspace{-1.2em}\textbf{Output:} Coreset $\mathbb{S}$ of size $M=\lceil \beta N\rceil$, where $M<<N$
\For{$t=1$ to $T$}
    \State $z^i \leftarrow f_\theta(x^i)$
    \State $\bm{\alpha^{i}} \leftarrow \text{ReLU}(\bm{z^{i}}) + \bm{1}$ 
    \State Obtain confidence score $S_{t}^{i} \leftarrow ||\boldsymbol{\alpha^i} ||_2$ 
    \State Track prediction correctness and calculate the cumulative number of forgetting events $R_t^i$ \Comment{Based on Eq.~\eqref{eq:R_t}}
    \State Get forgetting frequency $F_t^i$ \Comment{Based on Eq.~\eqref{eq:F_t}}
\EndFor
\State Update $V_t^i$ over a sliding window of size $K$ \Comment{Based on Eq.~\eqref{eq:V_t}}
\State Compute $U^i$ for each sample \Comment{Based on Eq.~\eqref{eq:U}}
\State Sort samples by $U^i$ in descending order
\State Select top $M$ samples as coreset $\mathbb{S}$
\State \Return $\mathbb{S}$
\end{algorithmic}
\rule{\linewidth}{0.8pt} 
\vspace{-25pt} 
\end{wrapfigure}
The core idea of our proposed Dynamic Unreliable-Driven Coreset Selection (DUCS) strategy is to quantify the unreliability of training samples from two complementary perspectives, including an inward-looking perspective and a backward-looking perspective. To facilitate understanding, the complete pipeline is summarized in Algorithm~\ref{alg: DUCS}, and all related definitions follow those introduced in the main text. 

Specifically, during training, DUCS updates each sample’s unreliability indicators at every epoch based on its predictive behavior. For each sample, the inward-looking perspective captures the dynamic variation of its confidence, while the backward-looking perspective characterizes its learning instability by measuring the frequency with which the sample is forgotten during training. After all training epochs are completed, DUCS computes the confidence variance within a fixed sliding window and combines it with the accumulated forgetting frequency to obtain an overall unreliability score for each sample. Finally, DUCS ranks all samples according to this score and selects the top $\lceil \beta N\rceil$ samples to construct the coreset, ensuring that the samples most informative for refining the decision boundary are preserved.
% 我就添加了一句过度的

Regarding the inward-looking confidence, we note that based on the Dempster–Shafer Theory of Evidence, the Dirichlet distribution differs fundamentally from standard probability-based confidence derived from Softmax outputs. Specifically, it models a distribution over class probabilities and explicitly encodes both relative class likelihoods and absolute evidence strength. The Dirichlet parameters therefore reflect not only the predicted class proportions but also the strength of the supporting evidence, enabling more reliable and effective uncertainty estimation. Our experiments further demonstrate the superiority of this design choice, and we will include additional theoretical analysis and discussion in the next version.

\section{Experiment}
\vspace{-5pt}
To verify the contribution of different components of the proposed method, we first evaluate the impact of different window size settings, followed by an analysis of the effectiveness of the two proposed coreset selection metrics. Finally, we report the computational cost of each component at the end of this section.

\vspace{-5pt}
\subsection{Experiment Setup}
\vspace{-5pt}
To ensure a fair comparison, all experiments in the appendix follow the same setup as described in the main text. The experiments were implemented in PyTorch and conducted on a single NVIDIA RTX 3090 GPU. Unless otherwise specified, both the coreset and surrogate networks employed the ResNet-18 architecture, and the surrogate network was trained for 200 epochs across all datasets.

For clarity, we adopt the following notations in subsequent experiments: ``DUCS$^{\#}$" denotes the variant that uses only the inward-looking perspective, while ``DUCS$^*$" denotes the variant that uses only the backward-looking perspective. 
\begin{wrapfigure}{R}{0.7\textwidth} 
    \vspace{-25pt}
    \centering
    \begin{subfigure}[b]{0.485\linewidth} 
        \centering
        \includegraphics[width=\textwidth]{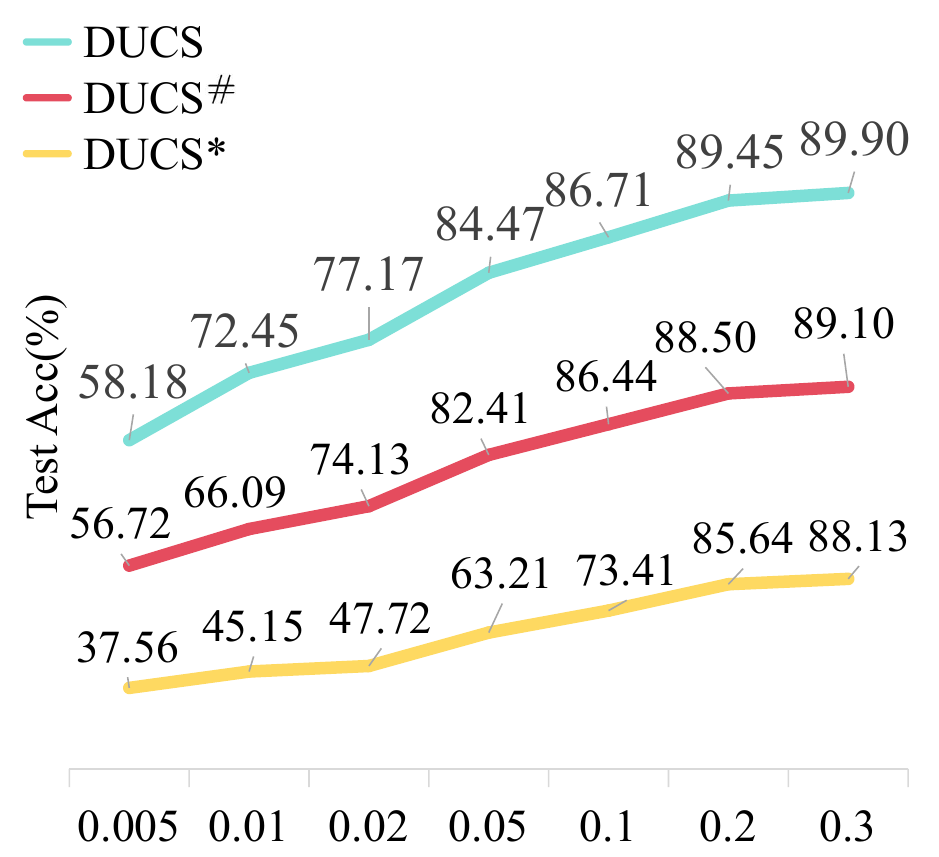}
        \caption{OrganSMNIST} 
        \label{fig: ap-s}
    \end{subfigure}
    \hfill
    \begin{subfigure}[b]{0.485\linewidth} 
        \centering
        \includegraphics[width=\textwidth]{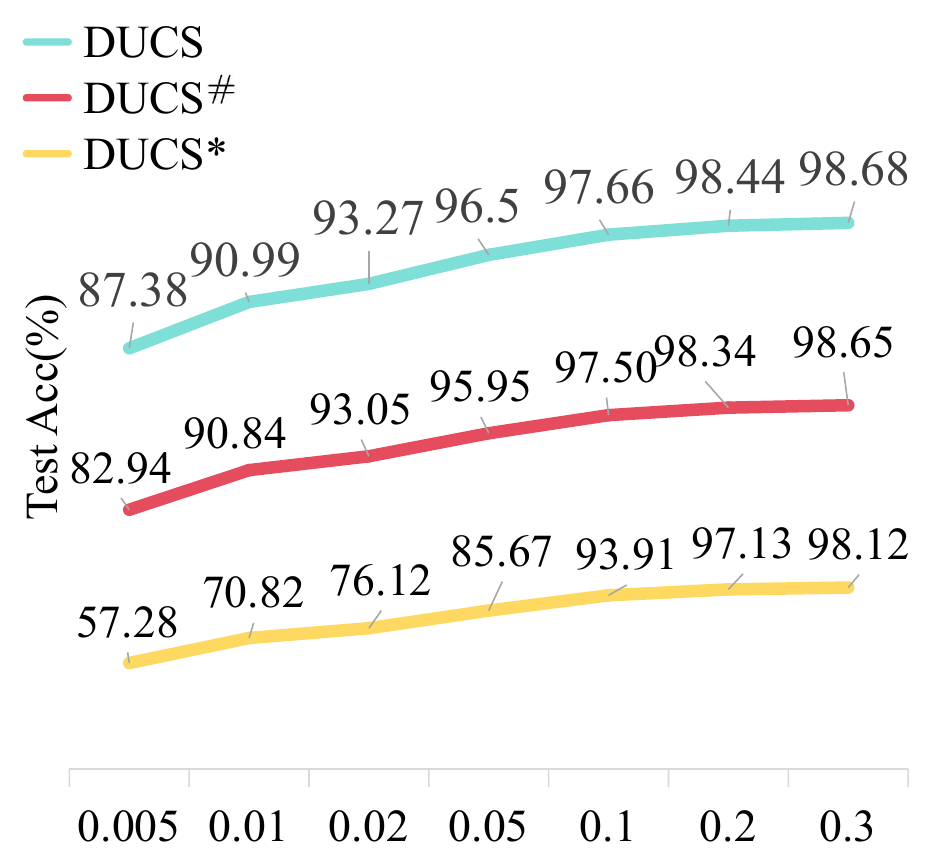}
        \caption{OrganAMNIST} 
        \label{fig: ap-a}
    \end{subfigure}
    \caption{Ablation study of the inward-looking and backward-looking perspectives. (a) and (b) denote the performance on different datasets.}
    \label{fig: ablation-ducs}
    \vspace{-20pt}
\end{wrapfigure}

\vspace{-5pt}
\subsection{Ablation Study about Different Selection Perspectives} 
\vspace{-5pt}
To further validate the effectiveness of the two proposed perspectives, we conduct experiments on the OrganAMNIST and OrganSMNIST datasets by enabling each perspective individually, and the results are shown in Figure~\ref{fig: ablation-ducs}. We can notice that using only the inward-looking or only the backward-looking perspective results in noticeable performance degradation compared to incorporating both perspectives, consistently across different coreset ratios. The results confirm that the inward- and backward-looking perspectives capture different and complementary aspects of sample unreliability.

\vspace{-5pt}
\subsection{Computational Cost.} 
\vspace{-5pt}
While inward scoring is relatively slower ($\sim$10x vs. backward), the absolute time overhead is minimal and negligible in the overall pipeline. As shown in Table~\ref{tab: compute time comparison}, the training time for DUCS remains significantly lower than that of full dataset training. Consequently, DUCS achieves superior model performance with a highly favorable trade-off between scoring cost and total training efficiency.
\begin{table}[!t]
\centering
\small
\caption{Compute time comparison on PathMNIST.}
\vspace{-10pt}
\label{tab: compute time comparison}
\resizebox{\columnwidth}{!}{
\begin{tabular}{c|ccc|c|c|c}
\toprule
$\beta$ & 2\% & 5\% & 10\% & Full Dataset & score computing time & selection time\\
\midrule
EVA & \multirow{2}{*}{123.03(3.36\%)} & \multirow{2}{*}{203.76(5.56\%)} & \multirow{2}{*}{323.92(8.85\%)} & \multirow{2}{*}{3660.53} & 17.00 & 1.00\\
DUCS & & & & & 23.09 & 1.00\\
\bottomrule
\end{tabular}
}
\vspace{-15pt}
\end{table}

\vspace{-3pt}
\subsection{Evaluation on Medical Datasets}
\vspace{-3pt}
To further validate the generalization capabilities of our proposed DUCS, we extended our evaluation to two additional medical datasets: \textbf{PathMNIST}~\cite{yang2023medmnist}
and \textbf{COVID-LDCT}~\cite{afshar2022human} (a high-resolution CT dataset). The quantitative results, summarized in Table~\ref{tab:path_covid_comparison}, demonstrate that DUCS outperforms state-of-the-art methods across various selection ratios ($\beta$). Notably, even at extremely low data fractions (e.g., $\beta = 2\%$ and $5\%$), our approach maintains robust accuracy. These results confirm that the superiority of our method extends beyond standard natural image benchmarks to diverse medical imaging modalities.
\begin{table}
    \centering
    \vspace{-15pt}
    \caption{Performance comparison on PathMNIST and COVID-LDCT datasets. The best and second-best results are marked in \textcolor{red!90!black}{red} and \textcolor{blue}{blue}, respectively.}
    \label{tab:path_covid_comparison}
    \vspace{-10pt} % 微调标题和表格的间距
    \resizebox{0.9\linewidth}{!}{
        {\setlength{\tabcolsep}{4pt} 
        \begin{tabular}{c|ccccc|ccccc}
        \toprule
        & \multicolumn{5}{c|}{\textbf{PathMNIST}} & \multicolumn{5}{c}{\textbf{COVID-LDCT}} \\
        $\beta$ & 2\% & 5\% & 10\% & 20\% & 30\% & 2\% & 5\% & 10\% & 20\% & 30\% \\
        \midrule
        Full dataset & \multicolumn{5}{c|}{99.38} & \multicolumn{5}{c}{98.97}  \\
        \midrule
        EVA & \textcolor{blue}{75.19} & \textcolor{blue}{77.90} & \textcolor{blue}{82.76} & 87.53 & \textcolor{blue}{91.10} & 54.63 & 60.32 & 61.82 & 71.83 & \textcolor{red!90!black}{96.57} \\
        Moderate-DS & 52.81 & 67.62 & 74.12 & 84.79 & 88.22 & \textcolor{blue}{66.48} & \textcolor{blue}{67.85} & \textcolor{blue}{69.77} & 77.52 & 80.47 \\
        TDDS & 73.40 & 77.59 & 82.51 & \textcolor{blue}{90.19} & 90.60 & 59.79 & 59.63 & 69.29 & \textcolor{blue}{89.10} & 92.05  \\
        \midrule
        \rowcolor{lightgray}
        \textbf{DUCS (Ours)} & \textcolor{red!90!black}{82.91} & \textcolor{red!90!black}{90.02} & \textcolor{red!90!black}{93.49} & \textcolor{red!90!black}{96.70} & \textcolor{red!90!black}{98.46} & \textcolor{red!90!black}{70.12} & \textcolor{red!90!black}{75.46} & \textcolor{red!90!black}{81.91} & \textcolor{red!90!black}{93.63} & \textcolor{blue}{95.13}  \\
        \bottomrule
        \end{tabular}
        }
    }
    \vspace{-25pt}
\end{table}

\begin{table}
    \centering
    \caption{Cross-architecture validation on PathMNIST and COVID-LDCT datasets: ResNet-18 $\to$ ViT-B/16.}
    \label{tab: VIT}
    \vspace{-10pt}
    \resizebox{0.9\linewidth}{!}{
        {\setlength{\tabcolsep}{3pt} 
        \begin{tabular}{c|ccccc|ccccc}
        \toprule
        & \multicolumn{5}{c|}{\textbf{PathMNIST}} & \multicolumn{5}{c}{\textbf{COVID-LDCT}} \\
        $\beta$ & 2\% & 5\% & 10\% & 20\% & 30\% & 2\% & 5\% & 10\% & 20\% & 30\% \\
        \midrule
        Moderate-DS & 54.18 & 56.36 & 64.50 & 68.95 & 75.92 & \textcolor{red!90!black}{59.15} & \textcolor{blue}{60.58} & \textcolor{red!90!black}{62.98} & 64.70 & \textcolor{blue}{68.54} \\
        TDDS & 54.87 & 75.88 & 81.18 & 84.93 & 84.30 & 50.51 & 51.54 & 52.64 & 50.45 & 53.80  \\
        EVA & \textcolor{blue}{75.85} & \textcolor{blue}{77.19} & \textcolor{blue}{83.38} & \textcolor{blue}{87.60} & \textcolor{blue}{91.14}  & 52.16 & 56.61 & 60.45 & 63.95 & \textcolor{blue}{67.51} \\
        \midrule
        \textbf{DUCS (Ours)} & \textcolor{red!90!black}{80.06} & \textcolor{red!90!black}{83.98} & \textcolor{red!90!black}{90.02} & \textcolor{red!90!black}{91.52} & \textcolor{red!90!black}{91.52} & \textcolor{blue}{57.92} & \textcolor{red!90!black}{61.82} & \textcolor{blue}{61.82} & \textcolor{red!90!black}{69.43} & \textcolor{red!90!black}{69.98} \\
        \bottomrule
        \end{tabular}
    }
    }
    \vspace{-20pt}
\end{table}

\vspace{-5pt}
\subsection{Generalizability to Architectures~(ViT).}
\vspace{-5pt}
To verify that the subsets selected by our method are not biased towards a specific network topology (e.g., convolutional neural networks), we further extended our evaluation to transformer-based architectures. Specifically, we conducted a cross-architecture validation where the coreset is selected using ResNet-18 and subsequently evaluated on a ViT-B/16 model. The results on the PathMNIST and COVID-LDCT datasets are reported in Table~\ref{tab: VIT}. As shown, DUCS consistently outperforms competing methods by a significant margin across various selection ratios. For instance, on the PathMNIST dataset at $\beta=10\%$, DUCS achieves an accuracy of 90.02\%, surpassing EVA by over 6\%. This proves that our approach possesses exceptional cross-architecture generalization capabilities.

\end{document}